\documentclass[a4paper, journal]{IEEEtran}
\usepackage{graphicx, color}
\usepackage{amsmath}
\usepackage{amssymb}
\usepackage{listings}
\usepackage{tabularx}
\usepackage{float}
\usepackage{multirow}

\definecolor{dkgreen}{rgb}{0,0.6,0}
\definecolor{gray}{rgb}{0.5,0.5,0.5}
\definecolor{mauve}{rgb}{0.58,0,0.82}

\begin{document}

\title{Developing efficient transfer learning strategies for robust scene recognition in mobile robotics using pre-trained convolutional neural networks}

\author{Hermann~Baumgartl,~\IEEEmembership{Student~Member,~IEEE,}
        and~Ricardo~Buettner,~\IEEEmembership{Member,~IEEE}
\thanks{This research is partly funded by the German Federal Ministry of Education and Research (no. 13FH176PX8, no. 13FH4I05IA, no. 13FH566KX). \emph{(Corresponding author: Hermann Baumgartl)}}

\thanks{The authors are with the Machine Learning Research Group Prof. Buettner, University of Bayreuth, 95447 Bayreuth, Germany: (email: hermann.baumgartl@hs-aalen.de; ricardo.buettner@uni-bayreuth.de. }
}

\markboth{Developing efficient transfer learning strategies for robust scene recognition}%
{Baumgartl \MakeLowercase{\textit{et al.}}: Development of efficient transfer learning strategies}

\maketitle

\begin{abstract}
We present four different robust transfer learning and data augmentation strategies for robust mobile scene recognition. By training three mobile-ready (EfficientNetB0, MobileNetV2, MobileNetV3) and two large-scale baseline (VGG16, ResNet50) convolutional neural network architectures on the widely available Event8, Scene15, Stanford40, and MIT67 datasets, we show the generalization ability of our transfer learning strategies. Furthermore, we tested the robustness of our transfer learning strategies under viewpoint and lighting changes using the KTH-Idol2 database. Also, the impact of inference optimization techniques on the general performance and the robustness under different transfer learning strategies is evaluated. Experimental results show that when employing transfer learning, Fine-Tuning in combination with extensive data augmentation improves the general accuracy and robustness in mobile scene recognition. We achieved state-of-the-art results using various baseline convolutional neural networks and showed the robustness against lighting and viewpoint changes in challenging mobile robot place recognition.
\end{abstract}


\IEEEpeerreviewmaketitle

\section{Introduction}

\IEEEPARstart{R}{obustness} is of critical importance in many mobile robot applications such as inertia measurement \cite{qin2018a}, visual odometry \cite{zhao2020a} or visual navigation \cite{wen2020a}, simultaneous localization and mapping (SLAM) \cite{cadena2016a, bescos2021a} and visual place recognition \cite{khaliq2020a, lowry2016a}.  Most vision systems are very well optimized for static environments with little to no variation in the image conditions \cite{bescos2021a}. However in real-world conditions, places may heavily change in visual appearance be it from moving objects \cite{suenderhauf2015a, bescos2021a}, occlusions of objects and landmarks \cite{arroyo2016a}, changing viewpoint conditions and lighting conditions \cite{mancini2017a} or combinations of these factors \cite{khaliq2020a,xie2020a}. Due to this high variety of place conditions and the high complexity of the places themselves \cite{arroyo2016a}, place recognition is one of the most difficult challenges \cite{khaliq2020a}.

Various sensors such as Sonar \cite{oore1997a, sanchez2012a}, Lidar \cite{zhangji2015a} or positioning systems like GPS \cite{agrawal2006a} have been used for robot localization and navigation. Place recognition, on the other hand, is mostly based on categorizing monocular images into predefined classes \cite{xie2020a}. In recent years deep convolutional neural networks (CNN) (re-)emerged as one of the prime computer vision algorithms for solving complex problems \cite{leCun2015a}. In the course of the return of CNNs transfer learning has emerged as a way to train large-scale CNNs on limited data. The intuition behind transfer learning is to utilize learned representation from another domain in the target domain \cite{pan2010a}, assuming that these representations are relevant to the target task to some degree \cite{li2018a}. In practice, this means that large CNN architectures are pre-trained on large datasets such as ImageNet before the network is fine-tuned on a much smaller dataset. The usage of transfer learning enables researchers to develop large architectures that would not be feasible for small datasets and then transfer them to their problem/domain.

\begin{figure}[ht]
  \centering
  \includegraphics[width=0.6\columnwidth]{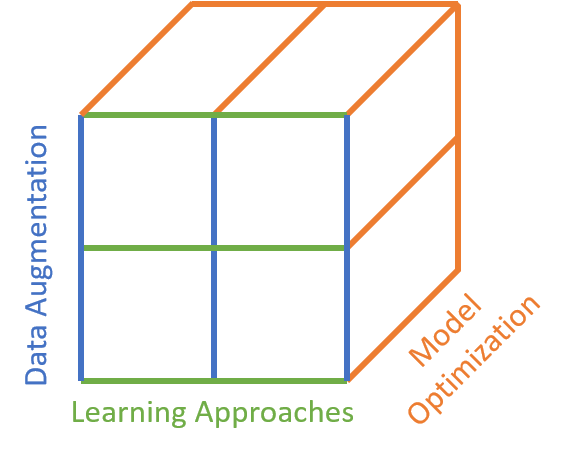}
  \caption{Dimensions of efficient transfer learning strategies for the use in mobile robotics. Efficient transfer strategies are developed by combining different learning approaches and data augmentation policies and are influenced by the model optimization techniques.}
  \label{fig:dimensions}
\end{figure}

Many indoor and outdoor place recognition algorithms based on CNNs have been proposed in the past years \cite{xie2020a}. While in general, the CNNs show superior performance even under heavy lighting and viewpoint changes \cite{mancini2017a, khaliq2020a}, many of these models come with high computational demands \cite{xie2020a, seong2020a} and are likely to run on high latency, especially when executed on low-power devices which is very crucial in mobile robotics \cite{zhao2020a, sandler2018a}. Many proposed models are based on large scale architectures such as VGG16 \cite{khaliq2020a} or ResNet \cite{seong2020a} which while they achieve good accuracies are not particularly optimized for mobile usage \cite{sandler2018a,howard2019a}. While different model optimization techniques such as pruning and quantization can be used to optimize models for inference in low-power devices, even optimized large-scale architectures often have higher latency when compared to architectures specially developed for mobile usage \cite{sandler2018a, howard2019a, mingxing2019a}.

Although various robust CNN models for place recognition have been proposed, many of these models are specifically designed for a single use case, not showing their transferability for other place recognition applications. Additionally, many robust models are also based on large-scale architectures with limited applicability on low-powered mobile robot platforms (see Table~\ref{tab:related}) \cite{xie2020a, seong2020a}. In this paper, we developed and tested different transfer learning strategies regarding their generalizability towards different place recognition datasets, their robustness against changing lighting and viewpoint conditions, and their inference speed on real-world mobile platforms. Furthermore, we evaluated the potential impact of inference speed optimization techniques on the robustness and effectiveness of our transfer learning strategies. We rigorously evaluated our transfer learning strategies using three well-established CNN architectures for mobile usage (MobileNetV2, MobileNetV3, EfficientNetB0) and compared them to the widely used large-scale architectures VGG16 and ResNet50. By evaluating their performance on the common place recognition datasets Event8 \cite{li2007a}, Scene15 \cite{lazebnik2006a}, Stanford40 \cite{yao2011a} and MIT67 \cite{quattoni2009a}, as well as evaluating their robustness using the KTH-Idol2 dataset \cite{luo2007a} we thoroughly test the predictive performance as well as the robustness using the developed transfer learning strategies. Furthermore, all models are also tested for their inference performance on low-power devices and how model optimizations like quantization influence the performance and robustness.

\noindent Our most important contributions are:\par
\begin{enumerate}
  \item We developed four different transfer learning and data augmentation strategies and benchmarked their effectiveness.
  \item Our transfer learning strategies achieve state-of-the-art results on multiple datasets while still maintaining a high transferability of the entire procedure.
  \item By conducting statistical tests on our results, we show that a combination of Fine-Tuning with aggressive data augmentation, a high level of generalization on different datasets, and good robustness can be achieved.
  \item We show the robustness of our strategies against severe changes in lighting and viewpoint conditions \cite{khaliq2020a}.
  \item We identified suitable mobile-ready baseline architectures to achieve state-of-the-art results in conjunction with our transfer learning strategies.
  \item By benchmarking different model inference optimization techniques, we show their effectiveness and impact on model recognition performance and robustness.
  \item We show that large-scale CNNs can be outperformed by mobile-ready architecture in terms of accuracy, robustness, and inference speed using transfer learning.
  \item Our approach is based upon widely available packages and can be easily adopted by other scholars.

\end{enumerate}

The paper is organized as follows: First, we give an overview of the related work, including a description of the current state of the art in performance for different datasets. Next, we provide the research methodology, with a description of the developed transfer learning strategies, the used deep learning architectures, datasets, model optimization techniques, and hardware accelerators. After that, we show the results of our transfer learning strategies in terms of generalization on different datasets, their robustness, and inference speed. We then discuss the results and their implications before concluding with limitations and suggestions for future research.

\section{Related Work}
Robust place recognition for mobile robots is one of the major challenges in mobile robotic research. In the following, the current state-of-the-art algorithms for the Event8, Scene15, and MIT67 datasets are described.

\renewcommand{\arraystretch}{1.15}
\begin{table}[ht]
\setlength{\tabcolsep}{4.8pt}
\centering
\caption{Performance of different machine learning based methods for robot place recognition (Adapted from \cite{xie2020a, seong2020a}).} \label{tab:related}
\begin{tabular}{llccc}
\noalign{\smallskip}\hline
\textbf{Method}           & \textbf{Architecture}   & \textbf{Event8} & \textbf{Scene15} & \textbf{MIT67} \\ \hline
DUCA \cite{khan2016a}            & AlexNet        & 98.70  & 94.50   & 71.80 \\
NNSD \cite{xie2020b}             & ResNet-152     & \underline{99.10}  & 94.70   & 85.40 \\
VS-CNN \cite{shi2019a}           & AlexNet        & 97.50  & \underline{97.65}   & 80.37 \\
CNN-NBNL \cite{mancini2017a} & VGG-16         & 97.04  & 95.12   & 82.49 \\
SDO \cite{cheng2018b}             & VGG-16         & -      & 95.88   & 86.76     \\
Multi-Scale CNNs \cite{herranz2016a} & VGG-16         & -      & 95.18   & 86.04 \\
ResNet-152-DFT+ \cite{ryu2018a} & ResNet-152     & -      & -       & 76.50 \\
Mix-CNN \cite{jiang2019a} & VGG-16     & -      & -       & 79.63 \\
FOSNet \cite{seong2020a}           & SE-ResNeXt-101 & -      & -       & \underline{90.37} \\
 \hline
\end{tabular}
\end{table}

The Non-Negative Sparse Decomposition model (NNSD) is a combination of a ResNet-152 as static feature extraction and a support vector machine for the final scene classification \cite{xie2020b}. It extracts patches of the scene at multiple scales, which are processed by a pre-trained ResNet-152 CNN. They extracted between 224 patches of 224x224 pixels and achieved accuracies between 85.40 and 99.10 percent \cite{seong2020a, xie2020b}.

The visually sensitive CNN model (VS-CNN) is a full AlexNet based CNN model \cite{shi2019a}. It uses a context-based saliency detection algorithm to construct visually sensitive region enhanced images, which helps the detection algorithm distinguish between visually important and unimportant regions. It uses 3 patches of 227x227 pixels and achieves accuracies between 80.37 and 97.65 percent \cite{seong2020a, shi2019a}.

FOSNet is a full convolutional neural network architecture for scene recognition. It uses two SE-ResNeXt-101 as baseline CNN architecture, the ObjectNet and PlacesNet, combined with a scene coherence loss. The scene coherence loss is used to train the PlacesNet and assumes that the label for a scene should not change over the entire image. The scene coherence loss is calculated by extracting multiple patches from a scene and predicting the scene label. They use 10 patches of 224x224 pixels and achieve an accuracy of 90.30 percent, which is currently the overall benchmark for the MIT67 dataset \cite{seong2020a}.

Ryu et al. \cite{ryu2018a} proposed a novel discrete Fourier transformation for ensemble networks (DFT+) in combination with different baseline CNN architectures as feature extractors. Using ResNet-152 they achieved an accuracy of 76.50 percent on the MIT67 dataset. However, in contrast to the current benchmark on the MIT67 dataset, they used a single patch of 224x224 pixels, which makes the ResNet-152-DFT+ the current benchmark for a single patch model with a small image \cite{seong2020a}.

The Mix-CNN uses deep patch representations from a local feature codebook which is a combined ImageNet1K and Places365 codebook of features encoded through a VGG16 CNN \cite{jiang2019a}. Like ResNet-152-DFT+ model, they use a single image patch. However, they use an image size of 448x448 pixels. With an accuracy of 76.50 percent, the Mix-CNN is the current benchmark for the MIT67 dataset with a single large image patch \cite{seong2020a}.

As the comparison and Table~\ref{tab:related} shows, all current state-of-the-algorithms are based upon large-scale CNN architectures. However, as shown in Table~\ref{tab:architectures} these architectures do come with a very high amount of parameters and floating-point operations (FLOPS).

\section{Training approaches for convolutional neural networks}
With the groundbreaking win of AlexNet on the ImageNet competition, CNNs (re-)emerged as one of the top methods for solving complex computer vision tasks \cite{leCun2015a}. Following the huge success of AlexNet, more sophisticated and powerful architectures like ResNet \cite{he2016a}, ResNeXt \cite{xie2017a}, Xception \cite{chollet2017a} or EfficientNets \cite{mingxing2019a} have been developed, increasing the performance yearly. CNNs have shown remarkable performance on different computer vision tasks, even outmatching the human performance in complex object detection tasks \cite{szegedy2015a}.

The major advantage of CNNs over non-deep learning computer vision approaches is the combination of feature extraction and final classification into one step. Traditional computer vision approaches consist of two separate stages: 1) extract hand-crafted features from the images, 2) using a (shallow) machine learning model to classify the image into predefined categories. Most shallow learning algorithms perform poorly on raw image data, and therefore both global and local image descriptors such as HOUP \cite{fazl_ersi2012a}, GIST \cite{oliva2001a}, SIFT \cite{lowe2004a}, SURF \cite{bay2006a}, CENTRIST \cite{wu2011a} and ORB \cite{rublee2011a} have been developed. These descriptors are used to extract features from the images by reflecting textural features in the image such as edges, bright and dark spots, and geometric properties, which can be used to categorize the image into one of the learned places \cite{fazl_ersi2012a}. However, due to the high variability of conditions under which the images are taken, hand-crafted descriptors often struggle with robustness \cite{suenderhauf2015a, khaliq2020a}. When using hand-crafted descriptors and shallow learning, the two-step process decouples the process of actual learning from the feature extraction. One has to guess the optimal feature extraction algorithm and parameter, often leading to sub-optimal results \cite{leCun2015a}.

On the other hand, CNN uses the backpropagation algorithm to optimize their convolutional layers, i.e., their feature extraction, with respect to the final classification task. Because of this coupling, CNNs can automatically learn a set of deep feature representations for high quantities of different objects. These deep features are one of the major reasons for the superior performance of CNNs and are also the key for the transfer learning capabilities of deep CNNs \cite{yosinski2014a}.

\subsection{Training approaches}
Unlike traditional computer vision approaches, massive amounts of training data are necessary to achieve good results when training deep CNNs. Large networks like ResNet50 often need 100,000 images and more training images when trained from scratch. Nevertheless, in most real-world applications, it is not possible to acquire such large datasets \cite{weiss2016a}. However, large CNNs can be trained on small datasets using transfer learning while still achieving very good performances. Transfer learning makes use of the deep feature representations of CNNs. The intuition behind transfer learning is that these deep feature representations present a general abstraction of visual features. These can be transferred from the original domain to the target domain \cite{pan2010a}.

In deep learning, transfer learning for CNNs is often used when training on small datasets. Typically an ImageNet pre-trained deep CNN is used as base network \cite{yosinski2014a}. By using the convolutional layers only, the networks can be used for either feature extraction or Fine-Tuning utilizing different training approaches.

For \emph{feature extraction} the workflow is very similar to classical computer vision approaches. The original final classifier of the base architecture is removed, and the convolutional layers are used to generate feature vectors from the input images. These vectors are then fed into other machine learning models to perform the final classification \cite{donahue2014a, wozniak2018a}.

\emph{Fine-Tuning} consists of multiple consecutive steps. In the first step, the convolutional layers stay untouched, and only the final classifier is replaced by a new randomly initialized classifier which will learn the classification for the transfer learning dataset. The weights in the convolutional base are frozen, and the new classifier is trained on the new dataset. By freezing the convolutional base weights, the pre-trained visual representation is preserved. In a second step, a portion of the convolutional layers is then unfrozen to allow the model to perform changes on the feature extraction for the new dataset \cite{simonyan2015a}. While this improves model performance in most cases, there is a risk of destroying co-adaption between neurons in the convolutional base, which is decreasing model performance \cite{yosinski2014a}.

With the steady increase in the utilization of CNNs, new training approaches have been proposed. The explicit inductive bias tries to preserve the initial knowledge of the network, using an adapted L$^{2}$ weight regularization \cite{li2018a}. The \emph{L$^{2}$-SP} regularization prohibits the network from changing the network's weights too far from initial pre-trained weights, allowing the network to adapt the convolutional base to the new dataset while still preserving initial knowledge from the ImageNet pre-training. The degree of regularization can be adjusted using the $\alpha$ and $\beta$ parameters, where higher values applied more penalization \cite{li2018a}. Unlike Fine-Tuning, the \emph{L$^{2}$-SP} training approach trains the entire network, which decreases the risk of destroying the co-adaption but increases the computational needs during training.

\subsection{Data augmentation}
The training of CNNs is often complemented using data augmentation. Data augmentation is an effective technique to increase model performance, used in many image-related tasks like image classification, object detection, and image segmentation \cite{cubuk2020a}. It tries to increase the diversity of a dataset by randomly alter the images using image processing. The idea behind data augmentation is to increase the diversity of the data and thereby teach the model a greater part of the natural image variety \cite{cubuk2020a}.

The data is usually augmented using multiple regular image processing techniques like rotation, image blur, image distortion, or image noise. By transforming the images, the dataset is enlarged artificially, and it represents a broader variety of images. For example, by simply rotating the image, the CNN is forced to learn rotation-invariant features, which increases the models' performance and the robustness of the model. Utilizing the rotation invariant feature, the network is capable of recognizing objects under similar conditions \cite{krizhevsky2012a, cubuk2020a}.

\begin{figure}
  \centering
  \includegraphics[width=1.0\columnwidth]{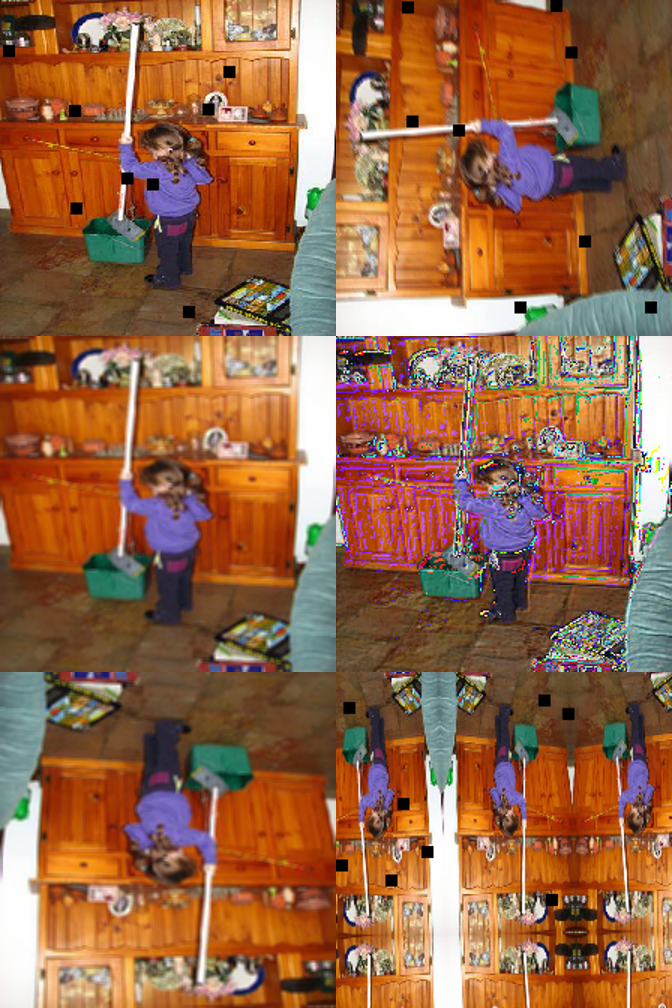}
  \caption{Example image from the Stanford40 dataset \cite{yao2011a} augmented using the \emph{ExtraAug} augmentation policy.}
  \label{fig:dataaug}
\end{figure}

Data augmentation has been successfully used to various degrees. While standard data augmentation policies usually include rotation, horizontal and vertical flipping, as well as random cropping or blur. However, more extensive data augmentation policies have also been used successfully \cite{buslaev2020a}. While standard data augmentation often leads to good results, specialized and more extensive data augmentation policies like grid distortion, coarse dropout, or affine transformation can increase the performance of deep CNN models \cite{cubuk2020a, buslaev2020a}.

\section{Convolutional neural network architectures}
\subsection{MobileNetV2}
The MobileNetV2 architecture is one of the prime convolutional neural network architectures for use in low-power devices. It is optimized to provide low memory consumption and low computational cost during inference \cite{sandler2018a}. The main improvement of MobileNetV2 is the usage of the inverted-residuals-with-linear-bottleneck-blocks, where low-dimensional representations are taken as an input, expanded to a high-dimensional representation by a regular convolution with ReLU non-linearity, filtered by a depthwise separable convolution and propagated back to a low-dimensional representation by a linear convolution \cite{sandler2018a}. Due to these blocks, the architecture does not need to fully materialize large intermediate tensors during inference, which greatly reduces memory consumption \cite{sandler2018a}.

The second main building blocks of the MobileNet architecture are depthwise separable convolutions \cite{howard2017a}, a specialized type of convolutional layers. A depthwise separable convolution performs a standard depthwise convolution across all input channels before building the new feature space using a pointwise convolution. By combining the feature space using the pointwise convolution, significant reductions in terms of computational cost and memory consumption can be achieved \cite{chollet2017a}.

To adapt the network to different hardware capabilities, the size of the network can be adjusted using a width multiplier, which thins or widen the network at each layer \cite{howard2017a}. Another parameter used to control the network size is the size of the input images, where the standard input size of 224x224 pixels is used. Smaller input sizes such as 96x96 pixels reduce the networks' size and computational costs in exchange for lower classification accuracy \cite{howard2019a}.

\subsection{MobileNetV3}
The MobileNetV3 architecture added several improvements over its predecessor. First of all, it uses a platform-aware neural architecture search (NAS) to search for efficient and lightweight architectures automatically. This enabled to search within a vast space of potential architecture candidates \cite{howard2019a}. Secondly, it uses the NetAdapt algorithm to determine the optimal number of convolutional filters at each layer to optimize accuracy. Using this combination of platform-aware NAS, the NetAdapt algorithm, and several other improvements, like the redesign of computational expensive layers, the architecture achieves better accuracy and an overall lower inference run time than MobileNetV2 \cite{howard2019a}.

The MobileNetV3 architecture also incorporates several manual improvements. The bottleneck-blocks have been extended using squeeze-and-excitation as well as the hard-swish activation function. These two extensions improved the overall accuracy of the MobileNetV3 architecture. However, both improvements are based upon the Sigmoid function, which is computationally expensive \cite{howard2019a, mingxing2019a}. Therefore two different versions of the MobileNetV3 architecture have been proposed. The MobileNetV3-Large architecture incorporates a total amount of 20 convolutional Blocks and the squeeze-and-excitation layers and is intended for use on platforms with higher computational power. The MobileNetV3 architecture also has minimalistic versions, which do not include advanced blocks like squeeze-and-excitation, hard-swish, and 5x5 convolutions. It is intended for use on low-powered platforms but with a reduced accuracy over the non-minimalistic versions \cite{howard2019a}. Further optimization of the network size and performance is also possible using different input sizes, and width multipliers \cite{sandler2018a}.

\renewcommand{\arraystretch}{1.15}
\begin{table}[ht]
\caption{Comparison of different CNN architectures. Top-1 Acc: Top-1 Accuracy on the 2012 ILSVRC ImageNet validation set. Parameters: Number of parameters including the original top classifier. FLOPS: Number of FLOPS including the top classifier. FLOPS measured using TensorFlow 2.3.1.}
\label{tab:architectures}
\begin{tabular}{lcrr}
\hline\noalign{\smallskip}
\textbf{Model}     & \textbf{Top-1 Acc} & \textbf{Parameters} & \textbf{FLOPS} \\
\hline\noalign{\smallskip}
VGG16              & 71.3\%             & 138,357,544         & 30,960,211,824 \\
VGG19              & 71.3\%             & 143,667,240         & 39,285,112,688 \\
ResNet50           & 74.9\%             & 25,636,712          & 7,728,446,320  \\
ResNet101          & 76.4\%             & 44,707,176          & 15,158,006,640 \\
ResNet152          & 76.6\%             & 60,419,944          & 22,588,771,184 \\
ResNeXt50          & 77.7\%             & 25,097,128          & 8,462,872,432  \\
ResNeXt101         & 78.7\%             & 44,315,560          & 15,941,906,288 \\
InceptionV3        & 77.9\%             & 23,851,784          & 11,450,968,624 \\
Inception-ResNetV2 & 80.3\%             & 55,873,736          & 26,343,534,128 \\
DenseNet121        & 75.0\%             & 8,062,504           & 5,670,888,304  \\
DenseNet169        & 76.2\%             & 14,307,880          & 6,722,308,080  \\
DenseNet201        & 77.3\%             & 20,242,984          & 8,585,415,920  \\
MobileNet          & 70.4\%             & 4,253,864           & 1,137,536,880  \\
MobileNetV2        & 71.3\%             & 3,538,984           & 601,617,264    \\
MobileNetV3-Small  & 67.4\%             & 2,554,968           & 2,976,688      \\
MobileNetV3-Large  & 75.2\%             & 5,507,432           & 5,589,536      \\
NasNetMobile       & 74.3\%             & 5,326,716           & 1,137,074,224  \\
NasNetLarge        & 82.4\%             & 88,949,818          & 47,679,813,660 \\
Xception           & 79.0\%             & 22,910,480          & 16,727,767,192 \\
EfficientNetB0     & 77.1\%             & 5,330,571           & 782,930,120    \\
EfficientNetB1     & 79.1\%             & 7,856,239           & 1,392,211,344  \\
EfficientNetB2     & 80.1\%             & 9,177,569           & 2,010,841,828  \\
EfficientNetB3     & 81.6\%             & 12,320,535          & 3,694,597,088  \\
EfficientNetB4     & 82.9\%             & 19,466,823          & 8,872,904,696  \\
EfficientNetB5     & 83.6\%             & 30,562,527          & 20,705,022,400 \\
EfficientNetB6     & 84.0\%             & 43,265,143          & 38,426,664,584 \\
EfficientNetB7     & 84.3\%             & 66,658,687          & 76,001,176,176 \\
\hline\noalign{\smallskip}
\end{tabular}
\end{table}

\subsection{EfficientNets}
EfficientNets is another family of CNN architectures designed for different purposes. The architectures are scaled from EfficientNetB0, which serves as mobile-ready baseline architecture, up to EfficientNetB7, which is a very large architecture with state-of-the-art accuracy on the ImageNet dataset \cite{mingxing2019a}. The baseline architecture EfficientNetB0 is also built using NAS with optimization for both accuracy and FLOPS to obtain a mobile-ready and efficient baseline architecture. Like MobileNetV2 and MobileNetV3, the mobile inverted bottleneck with added squeeze-and-excitation optimization and the swish activation function \cite{sandler2018a, howard2019a, mingxing2019a}.

This baseline model can then be scaled using the compound scaling coefficient to adapt the size of the network to different needs. In contrast to previous scaling approaches, the compound scaling scaled width, depth, and resolution of the convolutional layers simultaneously \cite{mingxing2019a}.

Since both squeeze-and-excitation and the swish activation are computationally expensive and can lead to problems during the model quantization (see Section~\ref{sec:optimization}), Lite versions of the EfficientNet are available. These do not include the squeeze-and-excitation, and the swish activation is replaced using a ReLU6 function \cite{mingxing2019a}.

\subsection{VGG16}
\begin{figure}
  \centering
  \includegraphics[width=1.0\columnwidth]{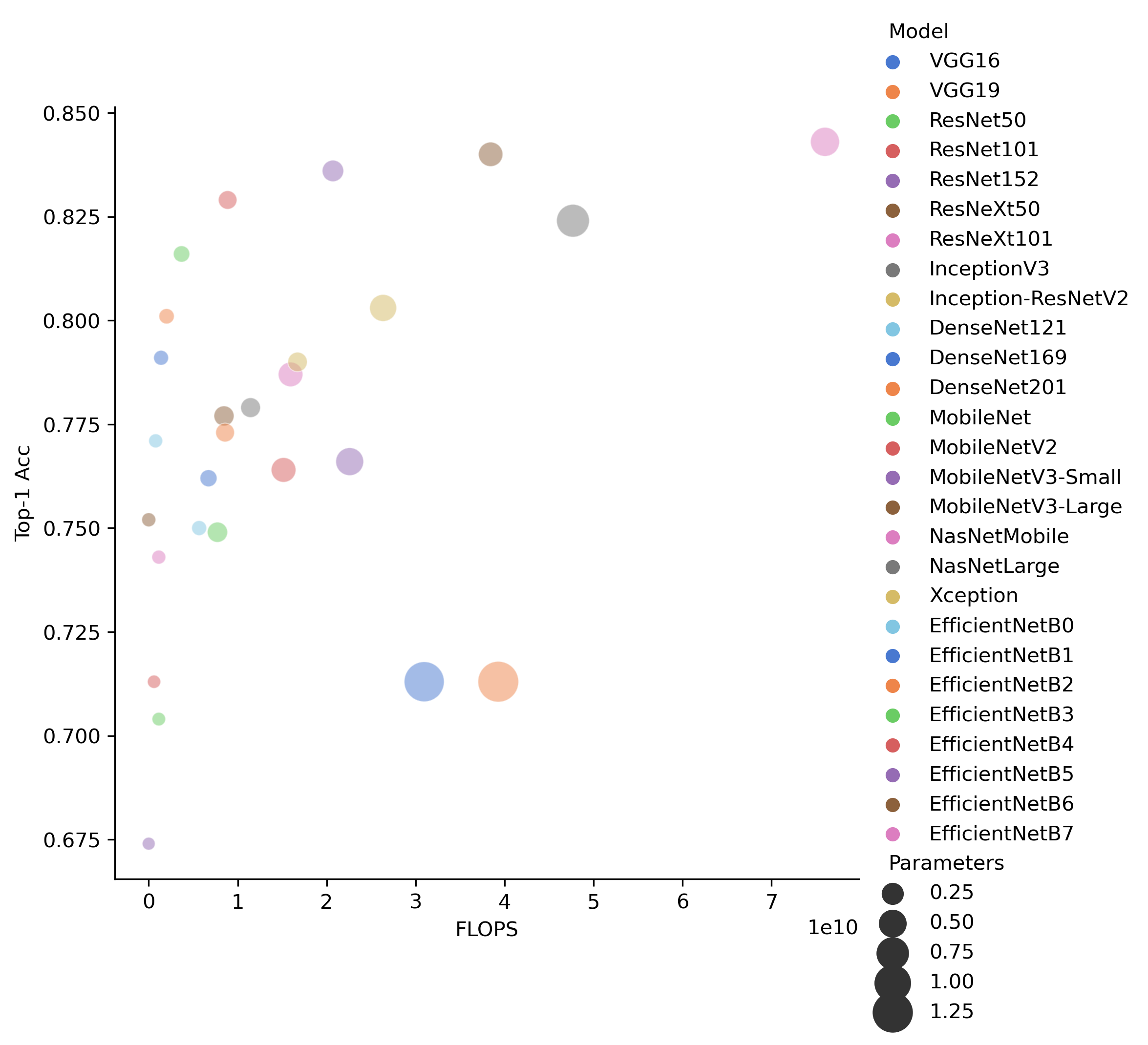}
  \caption{Comparison of different models in terms of Top-1 ImageNet accuracy, number of parameters and FLOPS. The size of the bubble denotes the number of parameters in million parameters}
  \label{fig:params}
\end{figure}

The VGG16 architecture is one of the first deep CNN architectures. It consists of 16 blocks of convolutional layers with pooling layers following \cite{simonyan2015a}. Many of the modern building blocks of deep CNN have been developed after the proposal of VGG16 and are therefore not included. VGG16 is considered as a classical architecture where convolutional layers are stacked upon to form a funnel shape where after each layer, the spatial dimension of the image is reduced while the depth of the tensor increases \cite{simonyan2015a}. While the 16 blocks of convolutional layers are considered as a small network size today, it is one of the computational expensive architectures with 138.3M parameters (see also Fig.~\ref{fig:params}).

Over the years, VGG16 has been used as a basis for many transfer learning approaches \cite{khaliq2020a, zhou2018a}. Due to its simplicity, it is easy to use for Fine-Tuning while still providing good accuracy. However, in recent years the VGG16 architecture has been replaced by the ResNet architecture as the main backbone for transfer learning \cite{seong2020a}.

\subsection{ResNet}
The ResNet50 architecture was the first architecture to introduce residual connections. These residual (or skip) connections counteract the problem of vanishing gradients and is a foundation of deeper CNNs \cite{he2016a}.

ResNet50 is the baseline architecture of the ResNet family with 50 convolutional layers. Due to its good robustness against the vanishing gradient, it quickly became one of the most adopted architectures for transfer learning. Other versions of ResNet include 18, 152 layers, or even 1001 layers showing the network's potential. Although the ResNet50 architecture is very popular, it is a large and computationally expensive architecture.

\section{Strategies for inference optimization}
\label{sec:optimization}
\subsection{Optimization techniques for inference optimization of convolutional neural networks}

Since their re-emergence, CNNs massively increased in terms of network complexity and computational cost. The complexity of a CNN model is commonly measured by the number of model parameters or floating-point operations (FLOPS) \cite{mingxing2019a}. FLOPS are measured during the inference of the model, meaning how many floating-point operations have to be calculated by the processor. Inference refers the model exection on data to generate predictions.

Current state of the art architectures are often very complex and have high amounts of network parameters and therefore high computational costs \cite{howard2017a, sandler2018a}. Networks such as VGG16 (138.3M parameters) \cite{simonyan2015a}, Inception-V3 (23.8M parameters) \cite{szegedy2015a}, ResNet-50 (25.5M parameters) \cite{he2016a} or Xception (22.8M parameters) \cite{chollet2017a} exceed the hardware capabilities of most mobile platforms and are therefore only partially feasible for implementation in such devices.

Since computational resources and power are limited on mobile devices, CNNs can only be implemented using efficient, and lightweight architectures \cite{howard2019a}. To build lightweight and efficient networks, many different approaches, such as training small networks from scratch \cite{arnold2017a} or shrinking pre-trained networks to meet resource limitations \cite{ba2014a, cheng2018a, han2016b, wu2016a} arose. Training small networks from scratch has been successfully used to build efficient networks \cite{arnold2017a}, but architecture engineering and optimization steps require significant amounts of time \cite{iandola2016a, zoph2018a}. One of the main building blocks for modern deep CNN architecture are depthwise separable convolutions \cite{chollet2017a}, a specialized type of convolutional layers. A depthwise separable convolution performs a standard depthwise convolution across all input channels before building the new feature space using a pointwise convolution. By combining the feature space using the pointwise convolution, great reductions in terms of computational cost and memory consumption can be achieved \cite{chollet2017a}. To obtain small networks from pre-trained networks, approaches such as shrinking \cite{iandola2016a}, network compression \cite{ba2014a, cheng2018a} and pruning \cite{han2016b} have been used to reduce network size. In addition to the existing pre-trained networks, architectures built for implementation in mobile and embedded devices have been proposed in the literature. Networks like MobileNetV2 \cite{sandler2018a} and MobileNetV3 \cite{howard2019a}, EfficientNetB0 \cite{mingxing2019a}, NASNet \cite{zoph2018a} and ShuffleNet \cite{zhang2018a} provide architectures for applications with low computational resources available.

One of the most common optimization techniques is quantization. It aims to reduce the bit width of the values range of the weights in a neural network. For example, 32-bit floating-point weights are commonly optimized to 8-bit fixed point range \cite{yang2019a}. The quantization procedure can be applied during (quantization aware training) or after training (post-training quantization) to parameters such as weights or activations of a neural network \cite{zebin2019a, nagel2020a}. Post-training quantization is easy to use since it can be applied to any trained network without significant retraining or adaption in the network architecture. However, it often suffers from a small to medium loss in accuracy. The quantization aware training introduces a fake loss is implemented during the training process to simulate the loss in accuracy due to the quantization. Therefore the network can adjust its weights with respect to the quantization \cite{nagel2020a}.


\subsection{Mobile deep neural network inference platforms}
In addition to the software optimization techniques for CNNs, various hardware accelerators are available. These accelerators are specialized to speed up the inference of neural networks on mobile platforms.

The Google Coral TPU accelerators are a series of mobile inference accelerators base on Google's Tensor Processing Units (TPU). The TPU is a coprocessor that is optimized to speed up the inference of 8-bit neural networks. The TPU requires 8-bit fully quantized neural networks and can speed up inference massively \cite{cass2019a}.

The Intel Neural Computer Stick is a USB accelerator for inference optimization on Windows and Linux platforms. It uses models converted into the OpenVINO Intermediate Representation format using FP16. The models are optimized for inference on a Myriad X Vision Processing Unit (VPU) which is specialized for the inference of neural networks \cite{ionica2015a, xu2017a}.

The Nvidia Jetson series is a family of GPU-based inference accelerators. The entry-level model is the Jetson Nano, which features a small energy-efficient Nvidia Tegra GPU on a Raspberry Pi-like board. Since it is a GPU-based board, it does not necessarily require any specialized optimization of the models. However, like all Nvidia GPUs, it greatly benefits from converting the models into a TensorRT model representation which is optimized to use a specific GPU most efficiently \cite{cass2020a}.

\section{Methodology}
In this section, we describe the used deep learning methods and transfer learning strategies. To rigorously evaluate our transfer learning strategies, we follow the specific machine learning guidelines \cite{leCun2015a, arlot2010a, yosinski2014a}.

\subsection{Training approaches}
To evaluate the effectiveness of different transfer learning strategies and their impact on model robustness, we used two different training approaches combined with two different data augmentation policies. All combinations of training and data augmentation policies have been evaluated for their baseline accuracy after training. To assess the impact of different inference optimization techniques on our transfer learning strategies, we also tested the impact on accuracy and inference performance using four optimization schemes:
\begin{enumerate}
    \item post-training quantization of weights and activations (PTQ)
    \item quantization of weights and activation after quantization aware training (QAT)
    \item optimization for GPU accelerators using Nvidia TensorRT (TRT)
    \item optimization for Myriad VPU's using Intel OpenVino (Myriad)
\end{enumerate}

\begin{figure*}
  \centering
  \includegraphics[width=2.0\columnwidth]{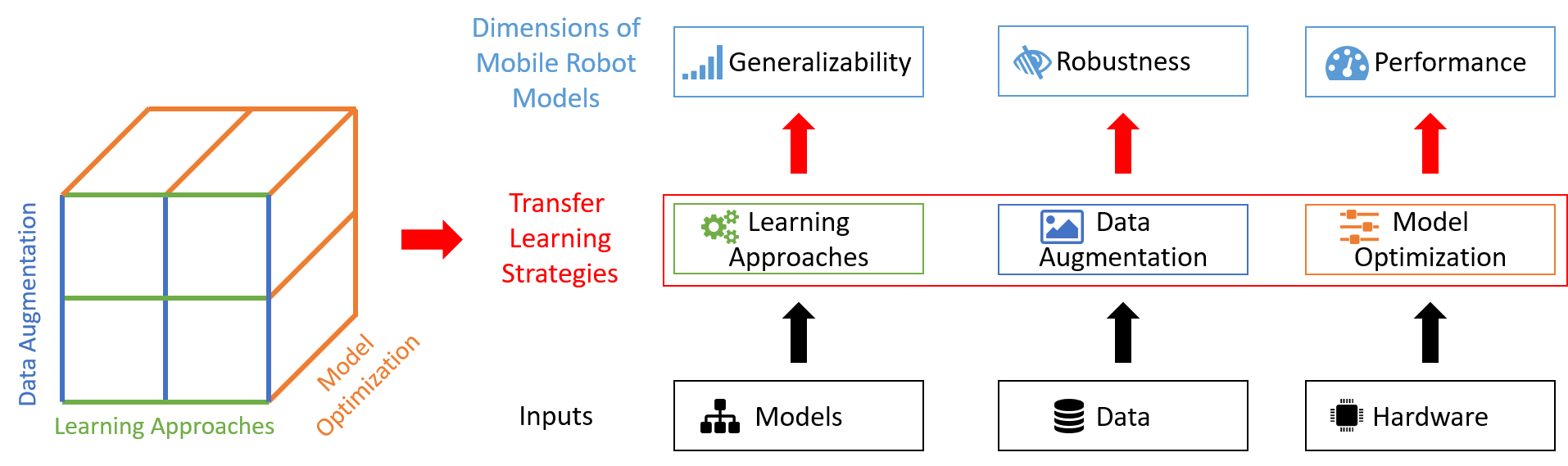}
  \caption{Methodology for the development and subsequent systematic benchmarking of our transfer learning strategies. The cube shows the possible dimensions of our transfer learning strategies. Mobile robot models can be measured in terms of generalizability, robustness, and performance. Therefore, our transfer learning strategies are measured in these three dimensions.}
  \label{fig:methodology}
\end{figure*}

As the first training procedure, we used the explicit inductive bias \emph{L$^{2}$-SP}, applied to all convolutional layers. Parameters of the explicit inductive bias where set to $\alpha$ = 0.1 and $\beta$ = 0.01. An Adam optimizer with an initial learning rate of 1e-5 was used for all training runs.

The second training approach is traditional Fine-Tuning. For the traditional Fine-Tuning we pre-trained the new classifier for 10 epochs before unfreezing the last two convolutional blocks of the respective models.

For training of the convolutional neural network, we used TensorFlow 2.3.1 \cite{abadi2016a}. The training ran on an Nvidia DGX A100 40GB (Ubuntu 20.04 LTS) for 100 epochs, with additional 50 epochs for the quantization aware training.

\subsection{Data augmentation policies}
In order to complement our transfer learning strategies, we implemented two different data augmentation policies. The first policy we implemented represents a relatively normal augmentation approach, including cropping and standard image alterations like rotations, flipping, and image blur (see Fig.~\ref{fig:normaug}).

\begin{figure}[t]
\begin{lstlisting}[language=Python]
def norm_aug(img_H, img_W, p=0.5):
    return albumentations.Compose([
        albumentations.PadIfNeeded(img_H, img_W, always_apply=True),
        albumentations.RandomCrop(img_H, img_W, always_apply=True),
        albumentations.RandomRotate90(p=0.5),
        albumentations.VerticalFlip(p=0.5),
        albumentations.HorizontalFlip(p=0.5),
        albumentations.Blur(blur_limit=3, p=0.5),
        ])
\end{lstlisting}
\caption{Python code for the NormAug augmentation policy, implemented as an albumentations pipeline. img\_H and img\_W represent the height and width for the resizing of the image. p is the probability to which an augmentation is applied.}\label{fig:normaug}
\end{figure}

In the second policy we implemented \emph{ExtraAug} the images are heavily augmented (see Fig.~\ref{fig:extraaug}). It represents an aggressive data augmentation procedure that includes the standard augmentation policy and adds various types of blur, distortion and sharpening filters, and coarse dropout.

\begin{figure}[t]
\begin{lstlisting}[language=Python]
def extra_aug(img_H, img_W, p=0.5):
    return albumentations.Compose([
        albumentations.PadIfNeeded(img_H, img_W, always_apply=True),
        albumentations.RandomCrop(img_H, img_W, always_apply=True),
        albumentations.RandomRotate90(p=0.5),
        albumentations.Flip(p=0.5),
        albumentations.OneOf([
            albumentations.OneOf([
                albumentations.IAAAdditiveGaussianNoise(),
                albumentations.GaussNoise(),
            ], p=0.2),
            albumentations.OneOf([
                albumentations.OpticalDistortion(p=0.3),
                albumentations.GridDistortion(p=0.2),
                albumentations.ElasticTransform(p=0.2),
            ]),
            albumentations.OneOf([
                albumentations.MotionBlur(p=.2),
                albumentations.MedianBlur(blur_limit=3, p=0.1),
                albumentations.Blur(blur_limit=3, p=0.1),
            ], p=0.2),
            albumentations.OneOf([
                albumentations.OpticalDistortion(p=0.3),
                albumentations.GridDistortion(p=.1),
                albumentations.IAAPiecewiseAffine(p=0.3),
            ], p=0.2),
            albumentations.OneOf([
                albumentations.IAASharpen(),
                albumentations.IAAEmboss(),
                albumentations.RandomBrightnessContrast(),            
            ], p=0.3),
        ]),
        albumentations.CoarseDropout(p=0.5)
    ], p=p)
\end{lstlisting}
\caption{Python code for the ExtraAug augmentation policy, implemented as an albumentations pipeline. img\_H and img\_W represent the height and width for the resizing of the image. p is the probability to which an augmentation is applied.}
\label{fig:extraaug}
\end{figure}

Before the data augmentation, all images we resized to 224x224 pixels. Since the pre-trained networks provided by TensorFlow were trained using different pixel value normalization, we used the preprocessing function provided with each network.

\subsection{Training architectures}
For the comparison of our transfer learning strategies used three common mobile-ready CNN architectures: MobileNetV2, MobileNetV3-Large (Minimalistic), and EfficientNetB0-Lite. We additionally tested the performance using the ResNet50 and VGG16 architecture to provide a benchmark. All architectures have been trained using the same train and test data to ensure comparability of the results.

In order to ensure full compatibility with the various inference optimization techniques and hardware accelerators, some changes in the CNN architectures are necessary. For the EfficientNetB0 architecture, we used the official EfficientNetB0-Lite architecture, which is specially optimized for good quantization performance on all hardware platforms. In contrast to the regular EfficienNetB0 architecture, the following changes are made: 1) the squeeze-and-excitation blocks are removed since they are not very well supported on some hardware accelerators, and 2) all swish activations have been replaced by ReLU6 for easier post-training quantization \cite{mingxing2019a}. Following the approach of the EfficientNet-Lite architectures we also replaced the swish activation function of the MobileNetV3-Large (minimalistic) models using ReLU6 for better quantization support.

As hyperparameters, we used the standard parameters for all architectures. The input size was fixed to 224x224 pixels. The width multiplier was set to 1 for all MobileNet architectures. All networks where initialized using the ImageNet weights provided by TensorFlow. As the last layer of the CNN networks, we used a global average pooling 2D layer followed by the new classifier. All new classifier consisted of a fully connected layer with 256 units, L2 regularization and ReLU activation, follower by a Dropout layer with 0.5 dropout probability, and the final fully-connected layer with a SoftMax activation (see Fig.~\ref{fig:classifier})
.
\begin{figure*}[t]
\begin{lstlisting}[language=Python]
x = layers.Dense(256, name='Dense_1', kernel_regularizer=l2(), activation='relu', input_shape=base_model.output_shape[1:])(base_model.output)
x = layers.Dropout(0.5, name='Dropout_2')(x)
x = layers.Dense(output_classes, name='output_D')(x)
output = layers.Activation('softmax', dtype='float32', name='output_S')(x)
\end{lstlisting}
\caption{Python code for the final classifier applied to every base CNN network, implemented in the Keras functional API}\label{fig:classifier}
\end{figure*}

\subsection{Benchmark datasets}
\label{sec:datasets}
In order to provide a highly comparable benchmark of our transfer learning strategies, we used multiple established indoor and outdoor place recognition dataset. To assess the overall performance of the models and their adaptability towards different scenarios we used the: Event8 \cite{li2007a}, Scene15 \cite{lazebnik2006a}, Stanford40 \cite{yao2011a} and MIT67 \cite{quattoni2009a} datasets.

\renewcommand{\arraystretch}{1.15}
\begin{table*}[ht]
\centering
\caption{Overview of the used datasets.}
\label{tab:datasets}
\begin{tabularx}{\textwidth}{lccX}
\hline\noalign{\smallskip}
\textbf{Dataset}    & \textbf{Number of images} & \textbf{Number of classes} & \textbf{Contents}                                                                                  \\
\hline\noalign{\smallskip}
Event8     & 1,572             & 8                 & Sport and event scenes e.g. badmitton, rowing, sailing, snowboarding                      \\
Scene15    & 4,485             & 15                & Indoor and outdoor places e.g. kitchen, mountain, forest, office                          \\
Stanford40 & 9,532             & 40                & Humans performing actions e.g. applauding, cooking, running, playing guitar, fixing a car \\
MIT67      & 15,620            & 67                & Indoor places e.g. airport, bakery, gym, concert hall, winecellar    \\
KTH-Idol2   & 14,893            & 5                 & Indoor places from two robot platforms: One/two person office, corridor, kitchen, printer area \\
\hline\noalign{\smallskip}
\end{tabularx}
\end{table*}

The Event8 dataset consists of 1,572 color images showing 8 classes of various sport and event scenes. The number of images per class ranges from 137 (bocce) to 250 (rowing) and contains various image conditions like cluttered and diverse image backgrounds, within object variations in size and varying poses and viewpoints \cite{li2007a}.
The Scene15 dataset contains 4,485 grayscale images showing 15 classes of natural and indoor scenes ranging from classes like 'kitchen' to 'forest'  \cite{lazebnik2006a}.
The Stanford40 dataset consists of 9,532 color images of humans performing 40 different actions under varying poses, viewpoints and with occasional object occlusions \cite{yao2011a}.
The MIT67 dataset contains 15,620 color images of 67 indoor scenes with a large scene variation, including general appearance, viewpoint variations, or different lighting conditions \cite{quattoni2009a}.

For the performance evaluation of the models, we performance a train/test split on the datasets where the entire dataset is split into a training and a test subset. For the Stanford40 and MIT67 datasets, we used the train/test splits provided with the datasets. In the case of the Event8, Scene15, and KTH-Idol2 datasets, no train/test splits were provided. Therefore we performed our own train/test splitting. The evaluation is only done using the testing data. This data has not been shown to the model before evaluation. In this way, potential overfitting of the models could be identified \cite{arlot2010a}.

\begin{figure}
  \centering
  \includegraphics[width=1.0\columnwidth]{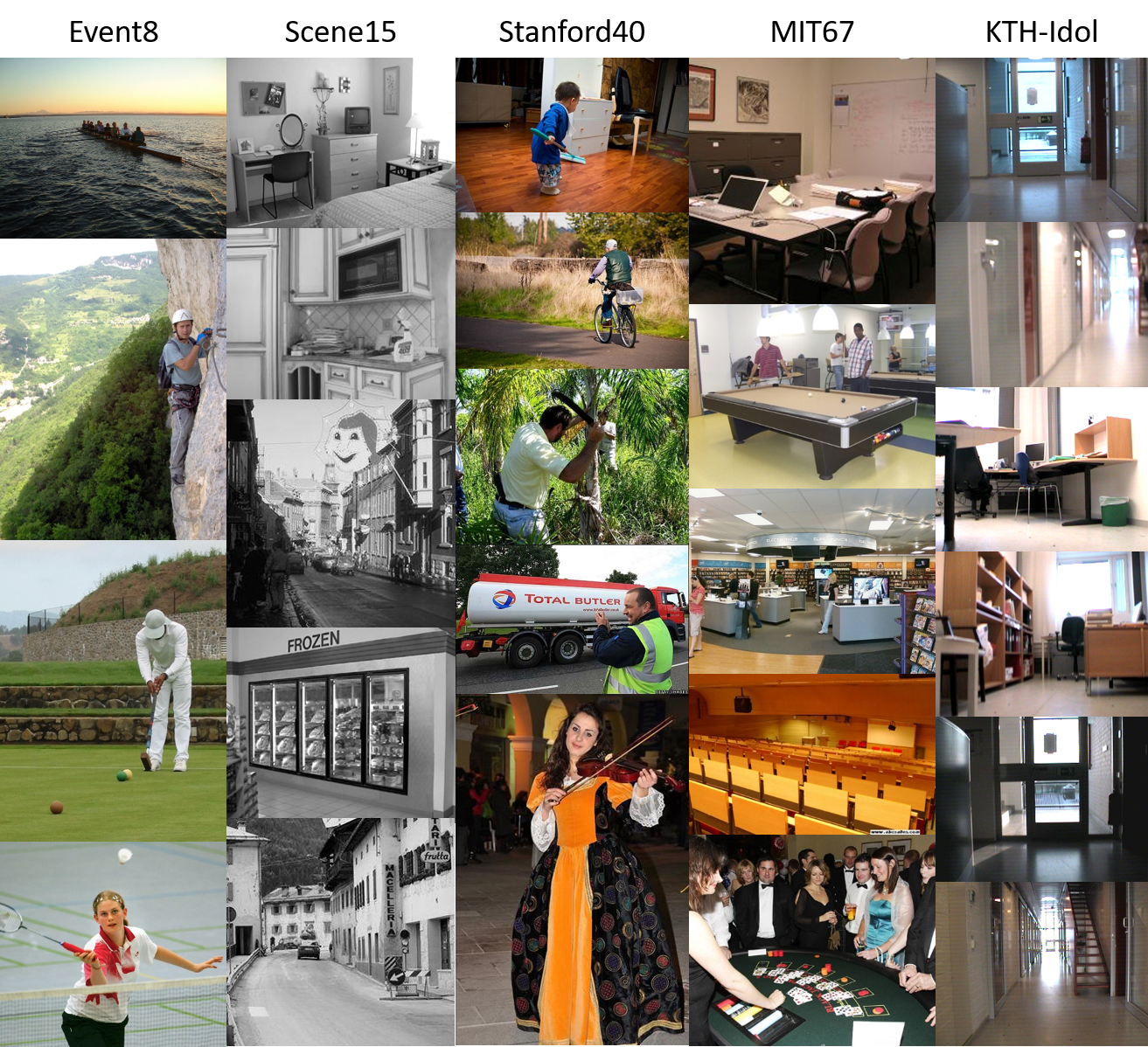}
  \caption{Sample images of used datasets.}
  \label{fig:examples}
\end{figure}

For the evaluation of the robustness of the transfer learning strategies towards changing lighting and viewpoint conditions, we used the KTH-Idol2 dataset \cite{luo2007a}. The KTH-Idol2 dataset is a well-established dataset consisting of 14,893 color images from 5 classes, showing places from an office environment. The images include two different mobile robots (Dumbo and Minnie) under two lighting conditions and various conditions like object occlusions, moved furniture, natural image blur, and moving persons. The images were acquired using a color camera mounted at different heights. For the Dumbo robot, the camera is 36 centimeters above the ground with a 13 degree tilt upwards, whereas the camera on Minnie is 98 centimeters above the floor with no tilt \cite{luo2007a}.

The images in the KTH-Idol2 dataset are tested in four different scenarios:
\begin{enumerate}
  \item Same robot for training and testing with same lighting conditions (SBSL)
  \item Same robot for training and testing with different lighting conditions (SBDL)
  \item Different robots for training and testing with same lighting conditions (DBSL)
  \item Different robots for training and testing with different lighting conditions (DBDL)
\end{enumerate}

\subsection{Model optimization and inference evaluation}
The trained models where optimized using four different optimization techniques. To optimize the models on CPU and TPU, we performed a full 8-bit post-training quantization of weights and activations (denoted PTQ) using TensorFlow Lite and the Coral Compiler 15.0.340273435. We additionally performed a quantization aware training in combination with the full 8-bit quantization of weights and activations (denoted QAT) using TensorFlow Lite.
Additionally, we converted the trained baseline models into efficient TensorRT representations (denoted TRT) for inference optimization on Nvidia GPU-powered platforms. We converted the models into two different representations using Float32 (FP32) and Float16 (FP16) as data types. All TensorRT conversions were performed on an Nvidia Jetson Nano 4GB following the official TensorRT conversion recommendations. The conversion was performed using Nvidia Polygraphy. Since the support to convert TensorFlow models to TensorRT representations does not yet fully support TensorFlow2 models, we converted our TensorFlow2 model into an intermediate ONNX version (OPSET 10) using the official TensorFlow2Onnx tool. All ONNX models have been sanitized using the Polygraphy Sanitizer to ensure compatibility with TensorRT. The maximum workspace size was set to 16.00 MB.


The models for the Intel Neural Compute Stick 2 (denoted Myriad) have been converted using the OpenVino Toolkit 2021.2. All models were directly converted from the saved TensorFlow2 models using a fixed batch size of 1 and FP16 floating-point range for compatibility with the Neural Compute Stick 2. No fixed preprocessing steps like fixed normalization were included in the Myriad models.

To test the execution speed for inference, we used two readily available and affordable mobile platforms. The inference performance was tested on a Raspberry Pi Model 4B 4GB with the Google Coral TPU USB Accelerator and an Intel Neural Compute Stick 2. Both accelerator sticks were connected to a USB 3.0 port of the Raspberry Pi. As the second inference evaluation platform, we use an Nvidia Jetson Nano 4GB. Both platforms were actively cooled using an external fan. As power supplies, the official Raspberry Pi power supply and a 5V/4A power supply were used, to avoid brown-out conditions.

For the inference benchmark, we used 32 images from the respective datasets. Before the inference test, we pre-loaded the images into the memory and performed 10 warm-up rounds to initialize the models and disable active hibernate states of the accelerators. The inference benchmark was repeated 10 times, and all timings were averaged across these 10 runs.

\section{Results}
In order to rigorously evaluate our transfer learning strategies, the analyses are divided into three separate evaluations. First of all, the generalization performance of our four transfer learning strategies is evaluated on the Event8, Scene15, Stanford40, and MIT67 datasets. The effect on the accuracy of the post-training quantization and the quantization aware-training are then compared to the baseline accuracy to assess the impact of these optimizations on the predictive performance of the models. Finally, the effect of the specialized optimization for the inference on the Jetson Nano and Intel Neural Compute Stick is evaluated.

The robustness of the transfer learning strategies, models, and optimization techniques is evaluated in the second part. We follow the same analysis procedure as for the general performance comparison by first evaluating the baseline performance before comparing the effect of the different optimizations on the models' performance. In contrast to the general performance comparison, we use the KTH-Idol2 dataset to assess the impact of the transfer learning and the optimizations on the robustness against changing lighting and viewpoint conditions.

We close our comparison by evaluating the inference performance of the different baseline and optimized models on real mobile computing platforms. Each model is benchmarked for its inference performance in terms of processing time for a batch of 32 images and its theoretical throughput of images.

\subsection{Generalization performance evaluation}
For the first analysis, we evaluated the baseline performance of the two training approaches (L2SP and Fine-Tuning) combined with our data augmentation procedures (NormAug and ExtraAug). All models are compared on the same testing sets. Since the class balance is not equal in most datasets, we report the balanced accuracy.

Table~\ref{tab:baseline_accuracy} shows that the EfficientNetB0-Lite architecture, in combination with Fine-Tuning and heavy data augmentation, outperforms the other models. Only for the Event8 dataset, does the MobileNetV2 architecture achieve a slightly better accuracy of 96.15 percent over 95.56 percent of the EfficientNetB0. In general, it can be seen that except for the VGG16 architecture, the range of the accuracy of the different models is very close together.

In almost all cases, the model training benefits from the more aggressive data augmentation in the ExtraAug scheme. Larger differences can also be seen when comparing the results of the L2SP/NormAug strategy to other strategies on the Stanford40 and MIT67 datasets. These datasets have significantly more classes than Scene15 and Event8. It, therefore, can be concluded that larger datasets with more classes benefit from a stronger data augmentation in all scenarios.

Furthermore, our results show that although the explicit inductive bias L2SP should preserve the co-adaption between layers and potentially fits the entire network, the models do not benefit from it in terms of balanced accuracy. In all scenarios in Table~\ref{tab:baseline_accuracy} the traditional Fine-Tuning archives similar or better results than the explicit inductive bias. This is especially true for the MobileNetV3 when applied to a larger number of classes.

\renewcommand{\arraystretch}{1.15}
\begin{table}[ht]
\centering
\caption{Baseline balanced accuracy of the models without prior optimization. All values in percent.}
\label{tab:baseline_accuracy}
\begin{tabular}{ll|cc|cc}
                            &                & \multicolumn{2}{c|}{L2SP} & \multicolumn{2}{c}{Fine-Tuning} \\
                            &                & NormAug     & ExtraAug    & NormAug        & ExtraAug        \\ \hline
\multirow{5}{*}{\rotatebox[origin=b]{90}{Event8}}     & EfficientNetB0 & 93.27     & 94.21     & 95.25        & 95.56         \\
                            & MobileNetV2    & 95.41     & 96.10     & 94.35        & \underline{96.15}         \\
                            & MobileNetV3    & 94.31     & 95.22     & 95.17        & 94.94         \\
                            & ResNet50       & 95.62     & 94.83     & 94.85        & 96.07         \\
                            & VGG16          & 92.73     & 95.21     & 95.36        & 95.64         \\ \hline
\multirow{5}{*}{\rotatebox[origin=b]{90}{Scene15}}    & EfficientNetB0 & 93.73     & 94.77     & 94.28        & \underline{95.19}         \\
                            & MobileNetV2    & 94.01     & 93.91     & 93.42        & 94.14         \\
                            & MobileNetV3    & 90.10     & 91.94     & 94.08        & 93.51         \\
                            & ResNet50       & 91.92     & 93.30     & 94.23        & 94.77         \\
                            & VGG16          & 90.91     & 91.62     & 93.50        & 92.65         \\ \hline
\multirow{5}{*}{\rotatebox[origin=b]{90}{Stanford40}} & EfficientNetB0 & 71.42     & 73.41     & 75.56        & \underline{77.97}         \\
                            & MobileNetV2    & 71.46     & 75.07     & 72.95        & 75.62         \\
                            & MobileNetV3    & 65.90     & 69.31     & 70.27        & 73.05         \\
                            & ResNet50       & 73.85     & 76.62     & 74.67        & 76.81         \\
                            & VGG16          & 62.66     & 67.86     & 68.47        & 70.37         \\ \hline
\multirow{5}{*}{\rotatebox[origin=b]{90}{MIT67}}      & EfficientNetB0 & 70.69     & 73.02     & 77.13        & \underline{79.19}         \\
                            & MobileNetV2    & 73.86     & 75.72     & 75.54        & 77.27         \\
                            & MobileNetV3    & 67.02     & 69.47     & 74.08        & 75.66         \\
                            & ResNet50       & 74.84     & 77.52     & 72.60        & 76.68         \\
                            & VGG16          & 66.30     & 70.72     & 71.40        & 73.60         \\
                            \hline\noalign{\smallskip}
\end{tabular}
\end{table}

Table~\ref{tab:quantization_accuracy} shows the performance impact of the post-training quantization (PTQ) as well as the quantization aware training (QAT) on the different models. It is striking that for the large-class dataset Stanford40 and MIT67, the performance drops hugely when a post-training quantization for all weights and activations is applied to the L2SP/ExtraAug trained models before the quantization aware training. Especially for the EfficienNetB0 and MobileNetV2 architecture, the performance drops by 20 to 30\%.

The effect of a drop in accuracy when applying post-training quantization is especially prevalent in the MIT67 dataset with the ExtraAug augmentation. While the EfficientNetB0 remains a good performing network, it can be seen that the quantization performance benefits from the quantization aware training. In some cases, the quantization aware models even outperform the baseline models. This is likely due to the additional 50 epochs of training. Only the MobileNetV2 architecture sometimes shows a degraded performance after quantization aware training on small datasets (see Event8 and Scene15 Fine-Tuning in Table~\ref{tab:quantization_accuracy}).

\renewcommand{\arraystretch}{1.15}
\begin{table*}[ht]
\setlength{\tabcolsep}{3pt}
\centering
\caption{Comparison of the balanced accuracy of the baseline model, after post-training quantization (PTQ) and after quantization aware training (QAT). All values are in percent.}
\label{tab:quantization_accuracy}
\begin{tabular}{ll|ccc|ccc|ccc|ccc}
                            &                & \multicolumn{3}{c|}{L2SP NormAug} & \multicolumn{3}{c|}{L2SP ExtraAug}   & \multicolumn{3}{c|}{Fine-Tuning NormAug}      & \multicolumn{3}{c}{Fine-Tuning ExtraAug}              \\
                            &                & Baseline   & PTQ       & QAT      & Baseline         & PTQ     & QAT     & Baseline         & PTQ     & QAT              & Baseline         & PTQ              & QAT              \\ \hline
\multirow{5}{*}{\rotatebox[origin=b]{90}{Event8}}     & EfficientNetB0 & 93.27    & 92.26   & 94.63  & 94.21          & 94.21 & 94.63 & 95.25          & 94.26 & \underline{95.63} & 95.56          & 95.46          & 95.28          \\
                            & MobileNetV2    & 95.41    & 95.46   & 94.86  & 96.10          & 96.20 & 95.48 & 94.35          & 94.32 & 93.84          & 96.15          & \underline{96.61} & 95.18          \\
                            & MobileNetV3    & 94.31    & 94.26   & 91.43  & \underline{95.22} & 92.62 & 92.03 & 95.17          & 93.20 & 89.05          & 94.94          & 93.32          & 91.69          \\
                            & ResNet50       & 95.62    & 95.25   & 96.00  & 94.83          & 94.83 & 94.95 & 94.85          & 94.85 & 95.11          & 96.07          & 96.07          & \underline{96.16} \\
                            & VGG16          & 92.73    & 92.73   & 91.62  & 95.21          & 93.65 & 92.85 & 95.36          & 94.82 & 91.31          & 95.64          & \underline{96.22} & 94.28          \\ \hline
\multirow{5}{*}{\rotatebox[origin=b]{90}{Scene15}}    & EfficientNetB0 & 93.73    & 91.62   & 94.41  & 94.77          & 93.01 & 94.74 & 94.28          & 93.08 & 94.95          & 95.19          & 94.88          & \underline{95.23} \\
                            & MobileNetV2    & 94.01    & 92.24   & 93.01  & 93.91          & 92.66 & 93.91 & 93.42          & 93.69 & 91.03          & \underline{94.14} & 93.44          & 92.09          \\
                            & MobileNetV3    & 90.10    & 87.00   & 91.72  & 91.94          & 88.57 & 92.82 & \underline{94.08} & 83.61 & 90.05          & 93.51          & 84.27          & 88.69          \\
                            & ResNet50       & 91.92    & 90.73   & 93.16  & 93.30          & 93.03 & 93.90 & 94.23          & 94.51 & 94.77          & 94.77          & \underline{94.98} & 94.35          \\
                            & VGG16          & 90.91    & 88.94   & 92.19  & 91.62          & 91.06 & 92.27 & \underline{93.50} & 93.03 & 92.66          & 92.65          & 92.24          & 93.12          \\ \hline
\multirow{5}{*}{\rotatebox[origin=b]{90}{Stanford40}} & EfficientNetB0 & 71.42    & 70.94   & 72.98  & 73.41          & 72.96 & 74.73 & 75.56          & 59.72 & 74.02          & \underline{77.97} & 77.62          & 76.42          \\
                            & MobileNetV2    & 71.46    & 70.30   & 72.14  & 75.07          & 74.40 & 75.00 & 72.95          & 72.28 & 73.68          & 75.62          & \underline{75.79} & 75.76          \\
                            & MobileNetV3    & 65.90    & 64.53   & 67.69  & 69.31          & 45.27 & 70.45 & 70.27          & 56.52 & 70.36          & \underline{73.05} & 68.51          & 72.84          \\
                            & ResNet50       & 73.85    & 73.82   & 73.42  & 76.62          & 76.68 & 76.92 & 74.67          & 74.57 & 73.12          & 76.81          & \underline{77.34} & 75.80          \\
                            & VGG16          & 62.66    & 62.34   & 62.23  & 67.86          & 66.87 & 68.61 & 68.47          & 64.59 & 66.30          & \underline{70.37} & 68.35          & 69.41          \\ \hline
\multirow{5}{*}{\rotatebox[origin=b]{90}{MIT67}}      & EfficientNetB0 & 70.69    & 67.14   & 73.62  & 73.02          & 41.34 & 73.83 & 77.13          & 72.34 & 77.11          & \underline{79.19} & 64.96          & 78.11          \\
                            & MobileNetV2    & 73.86    & 69.66   & 73.90  & 75.72          & 55.39 & 76.69 & 75.54          & 74.67 & 75.48          & \underline{77.27} & 75.68          & 77.04          \\
                            & MobileNetV3    & 67.02    & 61.85   & 68.06  & 69.47          & 64.53 & 71.53 & 74.08          & 63.59 & 74.80          & \underline{75.66} & 67.11          & 74.63          \\
                            & ResNet50       & 74.84    & 71.81   & 73.02  & \underline{77.52} & 75.54 & 76.89 & 72.60          & 72.64 & 71.06          & 76.68          & 76.24          & 75.67          \\
                            & VGG16          & 66.30    & 64.62   & 66.67  & 70.72          & 57.90 & 72.62 & 71.40          & 70.77 & 69.90          & \underline{73.60} & 67.15          & 73.04          \\
                            \hline\noalign{\smallskip}
\end{tabular}
\end{table*}

Table~\ref{tab:optim_accuracy} shows the balanced accuracy of the different models and training approaches when optimized for inference. Based on the results from Table~\ref{tab:quantization_accuracy} we use the accuracy of the quantization aware models as baseline performance. As the results show, there are only minor differences in accuracy between the quantization aware models (QAT), the TensorRT models (TRT), and the Intel Myriad VPU model (Myriad).

However, it can be seen that the TensorRT and Myriad models tend to perform slightly better than the quantization aware models. In contrast to the quantization aware models, both TensorRT and Myriad models still use float values to store weights and activations. Especially for the MIT67 dataset, the TensorRT and Myriad models outperform the 8-bit integer quantized models.

As in the previous analysis, the Fine-Tuning/ExtraAug strategy still performs best on the majority of the models. Only when looking at Event8/MobileNetV3 and Stanford40/ResNet50, the explicit inductive bias outperforms Fine-Tuning on the fully optimized models.

It can also be seen that the potential disadvantage coming from the limited 8-bit integer range in contrast to the much larger float32/16 range can be almost neglected using quantization aware training. Only when using MobileNetV3 in combination with Fine-Tuning, the models tend to benefit from the increased value range of FP32 and FP16.

\renewcommand{\arraystretch}{1.15}
\setlength{\tabcolsep}{2.5pt}
\begin{table*}[ht]
\centering
\caption{Comparsion of the balanced accuracy of the optimizations for different inference accelerators, TRT32: TensorRT FP32, TRT16: TensorRT FP16, Myriad: OpenVino Myriad VPU FP16. All values in percent.}
\label{tab:optim_accuracy}
\begin{tabular}{ll|cccc|cccc|cccc|cccc}
                            &                & \multicolumn{4}{c|}{L2SP NormAug}                                         & \multicolumn{4}{c|}{L2SP ExtraAug}                                        & \multicolumn{4}{c|}{Fine-Tuning NormAug}                                  & \multicolumn{4}{c}{Fine-Tuning ExtraAug}                                 \\
                            &                & QAT              & TRT32            & TRT16            & Movidus          & QAT              & TRT32            & TRT16            & Movidus          & QAT              & TRT32            & TRT16            & Movidus          & QAT              & TRT32            & TRT16            & Movidus          \\ \hline
\multirow{5}{*}{\rotatebox[origin=b]{90}{Event8}}     & EfficientNetB0 & \underline{94.63} & 93.27          & 93.27          & 93.53          & \underline{94.63} & 94.21          & 94.21          & 93.99          & \underline{95.63} & 95.25          & 95.25          & 95.00          & 95.28          & 95.56          & 95.56          & \underline{95.78} \\
                            & MobileNetV2    & 94.86          & \underline{95.41} & \underline{95.41} & 95.08          & 95.48          & \underline{96.10} & \underline{96.10} & 95.37          & 93.84          & \underline{94.35} & \underline{94.35} & 93.92          & 95.18          & 96.15          & 96.15          & \underline{96.19} \\
                            & MobileNetV3    & 91.43          & \underline{94.31} & \underline{94.31} & \underline{94.31} & 92.03          & 95.22          & 95.22          & \underline{95.90} & 89.05          & \underline{95.17} & \underline{95.17} & \underline{95.17} & 91.69          & 94.94          & \underline{95.20} & 94.74          \\
                            & ResNet50       & \underline{96.00} & 95.62          & 95.62          & 95.62          & \underline{94.95} & 94.83          & 94.83          & 94.83          & \underline{95.11} & 94.85          & 94.85          & 94.85          & \underline{96.16} & 96.07          & 96.07          & 96.07          \\
                            & VGG16          & 91.62          & \underline{92.73} & \underline{92.73} & \underline{92.73} & 92.85          & \underline{95.21} & \underline{95.21} & \underline{95.21} & 91.31          & \underline{95.36} & \underline{95.36} & \underline{95.36} & 94.28          & \underline{95.64} & \underline{95.64} & \underline{95.64} \\ \hline
\multirow{5}{*}{\rotatebox[origin=b]{90}{Scene15}}    & EfficientNetB0 & \underline{94.41} & 93.73          & 93.73          & 92.65          & 94.74          & \underline{94.77} & 93.83          & 94.39          & \underline{94.95} & 94.28          & 94.28          & 94.22          & 95.23          & 95.19          & 95.06          & \underline{95.64} \\
                            & MobileNetV2    & 93.01          & 94.01          & \underline{94.08} & 93.57          & 93.91          & 93.79          & 93.79          & \underline{94.13} & 91.03          & \underline{93.42} & \underline{93.42} & 93.02          & 92.09          & 94.14          & 94.14          & \underline{94.29} \\
                            & MobileNetV3    & \underline{91.72} & 90.10          & 90.10          & 90.12          & \underline{92.82} & 91.94          & 91.94          & 91.37          & 90.05          & \underline{94.08} & 94.01          & 93.59          & 88.69          & 93.51          & 93.51          & \underline{93.98} \\
                            & ResNet50       & \underline{93.16} & 91.92          & 92.31          & 91.99          & \underline{93.90} & 93.30          & 93.03          & 93.34          & \underline{94.77} & 94.23          & 94.16          & 94.07          & 94.35          & \underline{94.77} & \underline{94.77} & 94.77          \\
                            & VGG16          & \underline{92.19} & 90.91          & 90.91          & 90.91          & \underline{92.27} & 91.62          & 91.62          & 91.70          & 92.66          & \underline{93.50} & \underline{93.50} & 93.37          & \underline{93.12} & 92.65          & 92.65          & 92.67          \\ \hline
\multirow{5}{*}{\rotatebox[origin=b]{90}{Stanford40}} & EfficientNetB0 & \underline{72.98} & 71.43          & 71.45          & 71.14          & \underline{74.73} & 73.40          & 73.42          & 73.29          & 74.02          & 75.54          & 75.59          & \underline{75.68} & 76.42          & 78.00          & \underline{78.00} & 77.93          \\
                            & MobileNetV2    & \underline{72.14} & 71.44          & 71.51          & 70.63          & 75.00          & 75.02          & \underline{75.05} & 74.32          & \underline{73.68} & 72.95          & 72.90          & 72.21          & \underline{75.76} & 75.64          & 75.71          & 75.11          \\
                            & MobileNetV3    & \underline{67.69} & 65.83          & 65.88          & 65.74          & \underline{70.45} & 69.33          & 69.41          & 69.11          & 70.36          & 70.27          & 70.28          & \underline{70.50} & 72.84          & \underline{73.10} & 73.08          & 72.92          \\
                            & ResNet50       & 73.42          & 73.85          & \underline{73.91} & 73.67          & \underline{76.92} & 76.65          & 76.70          & 76.73          & 73.12          & 74.63          & \underline{74.75} & 74.46          & 75.80          & \underline{76.81} & 76.77          & 76.76          \\
                            & VGG16          & 62.23          & \underline{62.66} & 62.10          & 62.63          & \underline{68.61} & 67.86          & 67.56          & 67.57          & 66.30          & \underline{68.46} & 68.42          & 68.34          & 69.41          & \underline{70.35} & 70.31          & 70.25          \\ \hline
\multirow{5}{*}{\rotatebox[origin=b]{90}{MIT67}}      & EfficientNetB0 & \underline{73.62} & 70.69          & 70.69          & 70.65          & \underline{73.83} & 73.02          & 72.95          & 73.05          & 77.11          & 77.13          & \underline{77.21} & 76.59          & 78.11          & 79.19          & \underline{79.19} & 78.89          \\
                            & MobileNetV2    & 73.90          & \underline{73.95} & 73.86          & 73.29          & \underline{76.69} & 75.72          & 75.94          & 74.14          & 75.48          & \underline{75.69} & 75.68          & 75.33          & 77.04          & 77.19          & 77.29          & \underline{78.33} \\
                            & MobileNetV3    & \underline{68.06} & 67.02          & 67.02          & 66.98          & \underline{71.53} & 69.38          & 69.53          & 69.03          & \underline{74.80} & 74.08          & 74.08          & 73.67          & 74.63          & 75.66          & 75.66          & \underline{75.74} \\
                            & ResNet50       & 73.02          & \underline{74.84} & \underline{74.84} & 74.83          & 76.89          & \underline{77.52} & \underline{77.52} & 77.21          & 71.06          & 72.53          & 72.60          & \underline{72.70} & 75.67          & 76.61          & 76.84          & \underline{77.42} \\
                            & VGG16          & \underline{66.67} & 66.30          & 66.08          & 66.06          & \underline{72.62} & 70.72          & 70.26          & 70.87          & 69.90          & 71.33          & 71.40          & \underline{71.57} & 73.04          & \underline{73.60} & 73.53          & 73.32          \\ \hline
\end{tabular}
\end{table*}

\begin{figure}[ht]
  \centering
  \includegraphics[width=1.0\columnwidth]{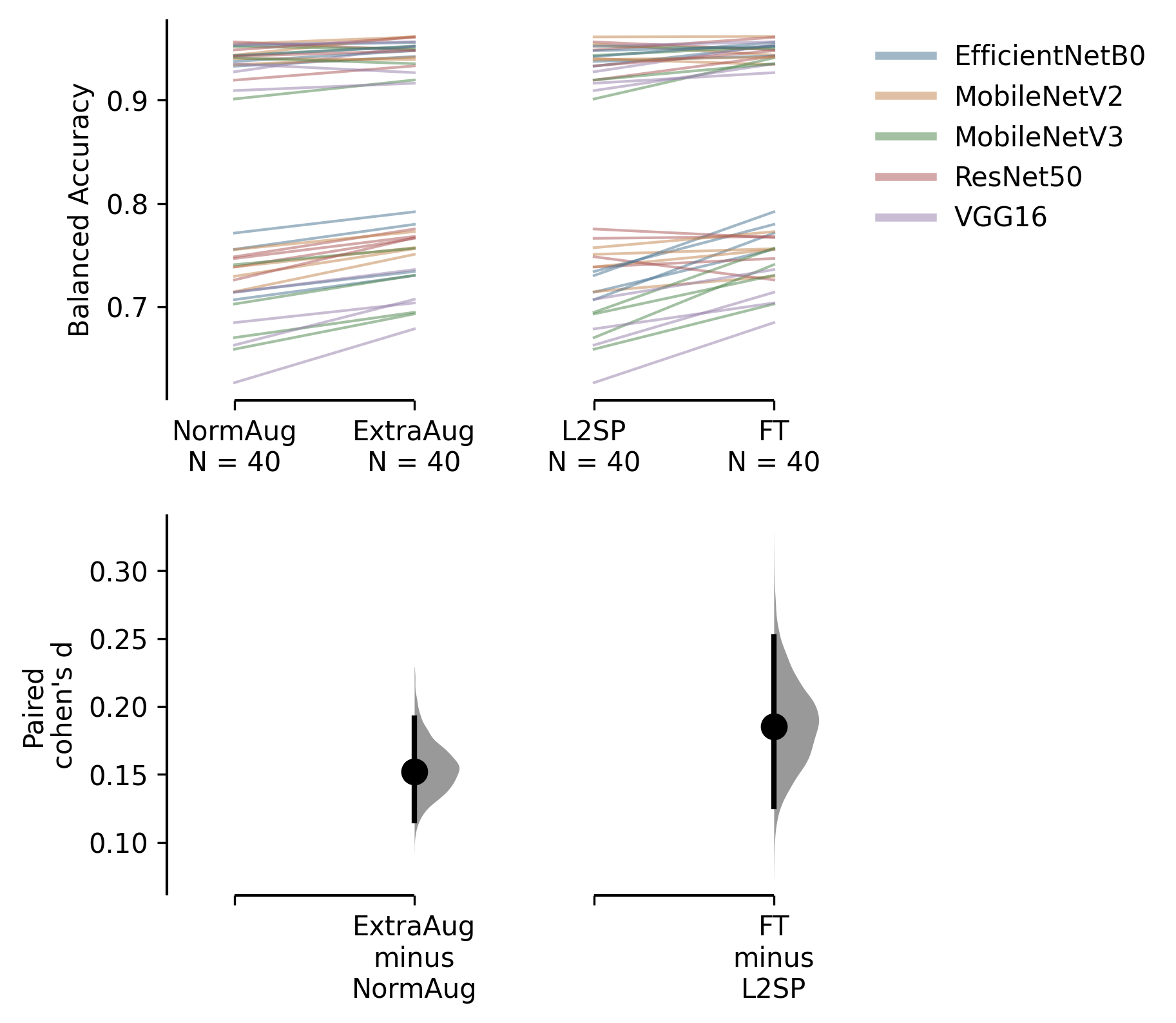}
  \caption{Slopegraph of the balance accuracy of the different transfer learning strategies.}
  \label{fig:dabest_strategies}
\end{figure}

The slopegraph in Fig.~\ref{fig:dabest_strategies} shows a comparison of the balanced accuracy of the two data augmentation and transfer learning strategies. When comparing the two different data augmentation schemes, the aggressive augmentation of the ExtraAug policy outperforms the standard NormAug policy in almost all cases. To further underpin this we performed a Wilcoxon signed-rank test (\(Z = 30.0\), \(p < 0.01\). The paired cohen's d effect size shows a small effect size of \(d = 0.15\) for the ExtraAug policy. In the comparison of the explicit inductive bias vs. the Fine-Tuning training approach, it can be seen that Fine-Tuning often outperforms L2SP. However, as Table~\ref{tab:baseline_accuracy} shows, this is also dependent on the data augmentation policy. With the exception of the MobileNetV3 model on the Event8 dataset, Fine-Tuning in combination with the ExtraAug augmentation policy outperforms L2SP with ExtraAug (Wilcoxon: \(Z = 74.0\), \(p < 0.01\); cohen's \(d = 0.18\)). All statistical tests have been performed using the dabest statistical package with 5,000 bootstrap resamples \cite{ho2019a}.

\subsection{Robustness evaluation}
In order to further assess the impact on the robustness of the different transfer learning strategies, we used the KTH-Idol2 dataset to evaluate the impact of changing lighting and viewpoint conditions. We followed a similar procedure as for the generalization performance by first analyzing the baseline performance, before doing an in-depth analysis of the impact of the different quantization and optimization techniques. All analyses are based on the four scenarios defined in the KTH-Idol2 dataset (see Section~\ref{sec:datasets}). All models were trained using either the Dumbo or Minnie robot platform using the 'cloudy' images \cite{luo2007a}. For the cloudy images, we performed a separate train/test split. All performance values for the 'same bot same lighting' condition are based on the test split only. The other conditions (SBDL, DBSL, DBDL) are entirely unseen for the model.

\renewcommand{\arraystretch}{1.15}
\begin{table*}[ht]
\centering
\caption{Baseline balanced accuracy for the KTH-Idol2 dataset. SBSL: Same Bot Same Lighting, DBSL: Different Bot Same Lighting, SBDL: Same Bot Different Lighting, DBDL: Different Bot Different Lighting. All values in percent.}
\label{tab:kth_baseline}
\begin{tabular}{ll|cccc|cccc|cccc|cccc}
                        &                & \multicolumn{4}{c|}{L2SP NormAug}     & \multicolumn{4}{c|}{L2SP ExtraAug}    & \multicolumn{4}{c|}{Fine-Tuning NormAug}      & \multicolumn{4}{c}{Fine-Tuning ExtraAug} \\
                        &                & SBSL    & SBDL    & DBSL    & DBDL    & SBSL    & SBDL    & DBSL    & DBDL    & SBSL    & SBDL    & DBSL    & DBDL             & SBSL     & SBDL     & DBSL     & DBDL     \\ \hline
\multirow{5}{*}{\rotatebox[origin=b]{90}{Dumbo}}  & EfficientNetB0 & 97.68 & 94.47 & 74.33 & 72.16 & 97.49 & 95.43 & 76.04 & 74.23 & 98.33 & 95.94 & 76.78 & 72.06          & 97.93  & 96.27  & 76.23  & 72.25  \\
                        & MobileNetV2    & 97.79 & 94.96 & 70.80 & 71.89 & 97.61 & 94.89 & 73.82 & 75.16 & 98.37 & 96.62 & 75.36 & 71.68          & 98.23  & 96.49  & 75.42  & 73.00  \\
                        & MobileNetV3    & 98.40 & 94.09 & 74.96 & 72.33 & 97.87 & 94.40 & 76.44 & 75.94 & 98.42 & 96.15 & 78.28 & 74.98          & 97.85  & 96.31  & 79.73  & 75.94  \\
                        & ResNet50       & 98.03 & 94.82 & 78.06 & 75.72 & 98.42 & \underline{96.71} & 82.17 & 78.67 & 98.37 & 95.00 & 75.54 & 73.81          & 98.46  & 94.95  & 79.14  & 75.84  \\
                        & VGG16          & 98.54 & 96.62 & 76.37 & 75.11 & 97.94 & 91.86 & 73.12 & 69.34 & 98.50 & 96.05 & 80.59 & 76.67          & 97.94  & 94.54  & 78.79  & 75.12  \\ \hline
\multirow{5}{*}{\rotatebox[origin=b]{90}{Minnie}} & EfficientNetB0 & 98.36 & 91.35 & 76.92 & 73.92 & 98.26 & 92.96 & 77.15 & 75.05 & 98.50 & 93.71 & 82.34 & 80.82          & 97.53  & 95.14  & 81.22  & 81.55  \\
                        & MobileNetV2    & 98.22 & 89.92 & 75.48 & 71.56 & 97.45 & 92.12 & 78.96 & 76.42 & 97.74 & 93.16 & 81.52 & 79.00          & 97.41  & 92.91  & 81.02  & 79.25  \\
                        & MobileNetV3    & 97.81 & 90.87 & 80.16 & 72.52 & 97.12 & 93.32 & 83.39 & 80.25 & 98.02 & 94.31 & \underline{86.40} & \underline{84.64} & 97.91  & 94.74  & 85.39  & 84.45  \\
                        & ResNet50       & 98.00 & 93.23 & 80.74 & 77.54 & 97.76 & 94.26 & 83.60 & 80.39 & 97.94 & 93.96 & 78.43 & 75.78          & 98.09  & 94.47  & 80.32  & 78.75  \\
                        & VGG16          & \underline{98.67} & 95.25 & 80.59 & 80.08 & 98.31 & 95.32 & 81.81 & 79.45 & 98.57 & 95.24 & 83.33 & 83.77          & 97.58  & 95.59  & 80.65  & 80.03  \\
                        \hline\noalign{\smallskip}
\end{tabular}
\end{table*}

Table~\ref{tab:kth_baseline} shows the baseline performance evaluation of our transfer learning strategies. When comparing the 'same bot same lighting' (SBSL) and 'same bot different lighting' (SBDL) conditions, it can be seen that all the models perform very well regardless of the transfer learning strategy. While for the Minnie robot platform, a performance drop can be seen using the NormAug data augmentation, the robustness increases when using the ExtraAug data augmentation. When comparing the conditions with different training and evaluation bots (DBSL / DBDL), the changes in viewpoint have a significantly larger impact than the changing lighting conditions. Especially when training with the Dumbo robot platform, a large performance drop from changing to the viewpoint of the Minnie bot can be seen. In the condition with the most changes 'different bot different lighting' (DBDL), there is only a little difference between the explicit inductive bias and the Fine-Tuning training approach. A much stronger effect can be observed when looking at the different data augmentation policies.

Like in the generalization performance analysis, the EfficientNetB0 performs well in all 'same bot' (SBSL / SBDL) conditions. However, when evaluating the 'different bot' (DBSL / DBDL) conditions, it can be seen that in terms of robustness, it is outperformed by MobileNetV3, which achieves results comparable or even outperforming to the much larger ResNet50. The VGG16 architecture achieves comparable results to the lightweight MobileNetV2/V3 and EfficientNetB0.

\begin{figure}[ht]
  \centering
  \includegraphics[width=1.0\columnwidth]{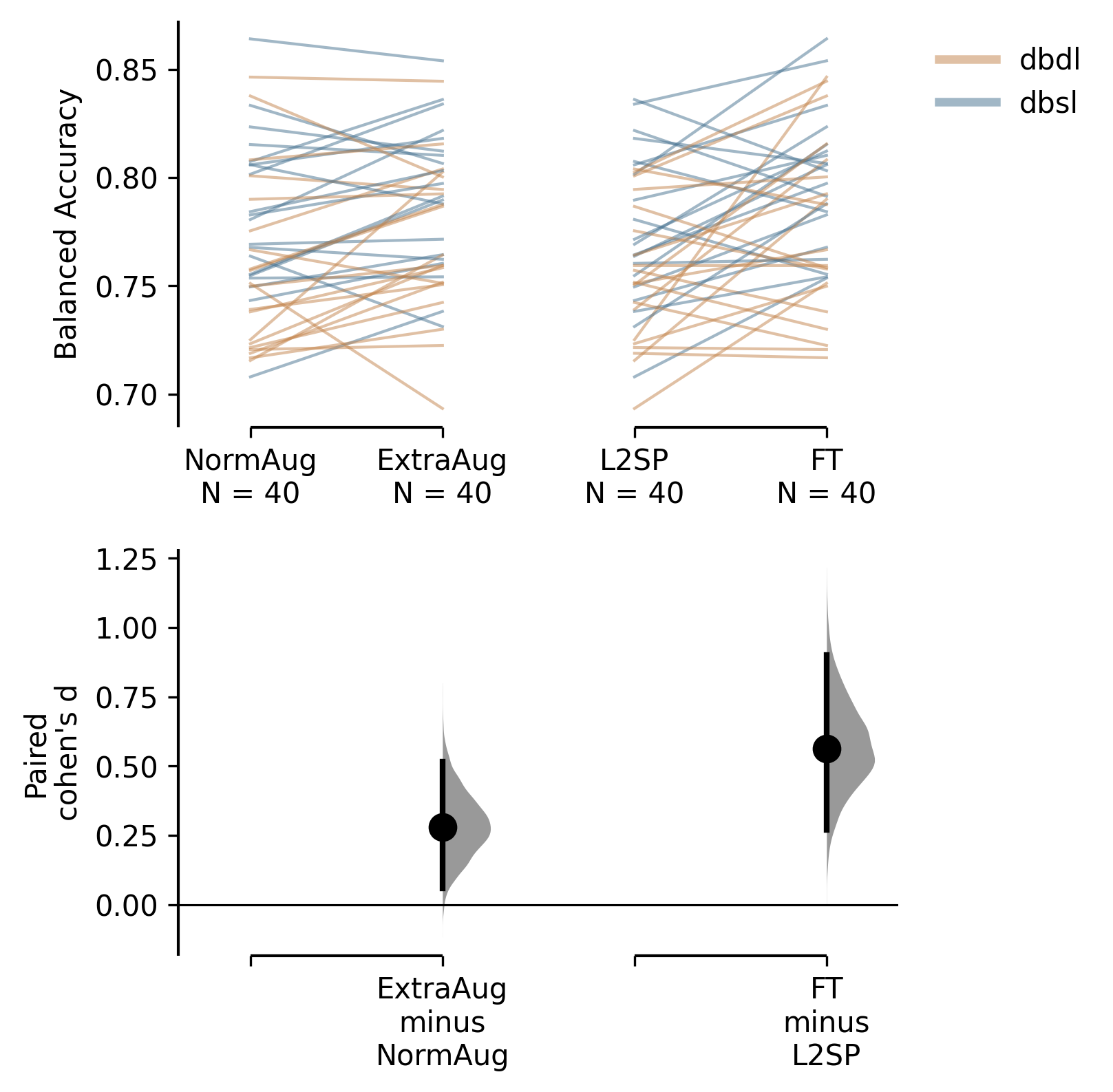}
  \caption{Slopegraph of the balance accuracy for the Different bot conditions of the KTH-Idol2 dataset.}
  \label{fig:dabest_kth}
\end{figure}

When comparing the impact of our transfer learning strategies in Table~\ref{tab:kth_baseline} and Fig.~\ref{fig:dabest_kth} similar results to the generalization evaluation can be seen. In general, the mobile-ready models benefit both from ExtraAug and Fine-Tuning. However, the large-scale baseline architectures VGG16 and ResNet50 do not always show better results when trained with Fine-Tuning and ExtraAug. The ResNet50 models show better results on 'different bot different lighting' scenarios when trained with L2SP, while the performance for VGG16 increases with the less aggressive NormAug data augmentation. This is especially prevalent when trained with the Dumbo images. However, when trained with the Minnie images, the mobile-ready CNN models show a better performance than the baseline architectures. While the effectiveness of the ExtraAug (Wilcoxon: \(Z = 215.0\), \(p < 0.01\); cohen's \(d = 0.27\)) is smaller compared to the generalization performance evaluation a larger effect from the Fine-Tuning training (Wilcoxon: \(Z =  173.0\), \(p < 0.01\); cohen's \(d = 0.56\)) can be observed.

\renewcommand{\arraystretch}{1.15}
\setlength{\tabcolsep}{3pt}
\begin{table*}[ht]
\centering
\caption{Balanced accuracy of the baseline model, after post-training quantization (PTQ) and after quantization aware training (QAT) on KTH-Idol2 with different bot conditions. All values are in percent.}
\label{tab:kth_quant}
\begin{tabular}{ll|ccc|ccc|ccc|ccc}
 &
   &
  \multicolumn{3}{c|}{L2SP NormAug} &
  \multicolumn{3}{c|}{L2SP ExtraAug} &
  \multicolumn{3}{c|}{Fine-Tuning NormAug} &
  \multicolumn{3}{c}{Fine-Tuning ExtraAug} \\
 &
   &
  \begin{tabular}[c]{@{}c@{}}Baseline\\ DBSL\end{tabular} &
  \begin{tabular}[c]{@{}c@{}}PTQ\\ DBSL\end{tabular} &
  \begin{tabular}[c]{@{}c@{}}QAT\\ DBSL\end{tabular} &
  \begin{tabular}[c]{@{}c@{}}Baseline\\ DBSL\end{tabular} &
  \begin{tabular}[c]{@{}c@{}}PTQ\\ DBSL\end{tabular} &
  \begin{tabular}[c]{@{}c@{}}QAT\\ DBSL\end{tabular} &
  \begin{tabular}[c]{@{}c@{}}Baseline\\ DBSL\end{tabular} &
  \begin{tabular}[c]{@{}c@{}}PTQ\\ DBSL\end{tabular} &
  \begin{tabular}[c]{@{}c@{}}QAT\\ DBSL\end{tabular} &
  \begin{tabular}[c]{@{}c@{}}Baseline\\ DBSL\end{tabular} &
  \begin{tabular}[c]{@{}c@{}}PTQ\\ DBSL\end{tabular} &
  \begin{tabular}[c]{@{}c@{}}QAT\\ DBSL\end{tabular} \\ \hline
\multirow{5}{*}{\rotatebox[origin=b]{90}{Dumbo}} &
  EfficientNetB0 &
  74.33 &
  72.51 &
  75.16 &
  76.04 &
  73.56 &
  75.39 &
  76.78 &
  75.84 &
  75.66 &
  76.23 &
  75.55 &
  74.54 \\
 &
  MobileNetV2 &
  70.80 &
  67.72 &
  71.20 &
  73.82 &
  69.74 &
  77.99 &
  75.36 &
  70.84 &
  71.33 &
  75.42 &
  71.48 &
  72.35 \\
 &
  MobileNetV3 &
  74.96 &
  63.25 &
  72.39 &
  76.44 &
  68.91 &
  74.62 &
  78.28 &
  58.38 &
  69.25 &
  79.73 &
  64.16 &
  67.87 \\
 &
  ResNet50 &
  78.06 &
  74.16 &
  79.58 &
  82.17 &
  82.38 &
  82.10 &
  75.54 &
  75.36 &
  79.16 &
  79.14 &
  79.22 &
  80.33 \\
 &
  VGG16 &
  76.37 &
  76.13 &
  79.10 &
  73.12 &
  73.25 &
  77.22 &
  80.59 &
  80.44 &
  80.16 &
  78.79 &
  79.11 &
  81.81 \\ \hline
\multirow{5}{*}{\rotatebox[origin=b]{90}{Minnie}} &
  EfficientNetB0 &
  76.92 &
  76.07 &
  75.91 &
  77.15 &
  78.04 &
  78.15 &
  82.34 &
  81.75 &
  77.64 &
  81.22 &
  81.07 &
  76.87 \\
 &
  MobileNetV2 &
  75.48 &
  76.36 &
  76.26 &
  78.96 &
  75.53 &
  74.71 &
  81.52 &
  81.35 &
  78.24 &
  81.02 &
  79.89 &
  78.07 \\
 &
  MobileNetV3 &
  80.16 &
  71.74 &
  77.01 &
  83.39 &
  79.25 &
  82.22 &
  86.40 &
  72.53 &
  80.09 &
  85.39 &
  70.42 &
  79.50 \\
 &
  ResNet50 &
  80.74 &
  80.52 &
  82.70 &
  83.60 &
  80.47 &
  80.96 &
  78.43 &
  77.82 &
  79.46 &
  80.32 &
  80.35 &
  81.04 \\
 &
  VGG16 &
  80.59 &
  79.75 &
  \underline{83.52} &
  81.81 &
  81.16 &
  77.16 &
  83.33 &
  83.14 &
  79.13 &
  80.65 &
  79.99 &
  82.57 \\ \hline\noalign{\smallskip}

 &
   &
  \multicolumn{3}{c|}{L2SP NormAug} &
  \multicolumn{3}{c|}{L2SP ExtraAug} &
  \multicolumn{3}{c|}{Fine-Tuning NormAug} &
  \multicolumn{3}{c}{Fine-Tuning ExtraAug} \\
 &
   &
  \begin{tabular}[c]{@{}c@{}}Baseline\\ DBDL\end{tabular} &
  \begin{tabular}[c]{@{}c@{}}PTQ\\ DBDL\end{tabular} &
  \begin{tabular}[c]{@{}c@{}}QAT\\ DBDL\end{tabular} &
  \begin{tabular}[c]{@{}c@{}}Baseline\\ DBDL\end{tabular} &
  \begin{tabular}[c]{@{}c@{}}PTQ\\ DBDL\end{tabular} &
  \begin{tabular}[c]{@{}c@{}}QAT\\ DBDL\end{tabular} &
  \begin{tabular}[c]{@{}c@{}}Baseline\\ DBDL\end{tabular} &
  \begin{tabular}[c]{@{}c@{}}PTQ\\ DBDL\end{tabular} &
  \begin{tabular}[c]{@{}c@{}}QAT\\ DBDL\end{tabular} &
  \begin{tabular}[c]{@{}c@{}}Baseline\\ DBDL\end{tabular} &
  \begin{tabular}[c]{@{}c@{}}PTQ\\ DBDL\end{tabular} &
  \begin{tabular}[c]{@{}c@{}}QAT\\ DBDL\end{tabular} \\ \hline
\multirow{5}{*}{\rotatebox[origin=b]{90}{Dumbo}} &
  EfficientNetB0 &
  72.16 &
  70.25 &
  74.12 &
  74.23 &
  72.77 &
  73.39 &
  72.06 &
  70.89 &
  71.46 &
  72.25 &
  70.99 &
  72.12 \\
 &
  MobileNetV2 &
  71.89 &
  69.47 &
  72.12 &
  75.16 &
  70.23 &
  78.24 &
  71.68 &
  69.04 &
  70.34 &
  73.00 &
  70.10 &
  71.13 \\
 &
  MobileNetV3 &
  72.33 &
  55.78 &
  68.84 &
  75.94 &
  65.67 &
  71.88 &
  74.98 &
  52.62 &
  68.00 &
  75.94 &
  59.63 &
  65.98 \\
 &
  ResNet50 &
  75.72 &
  73.25 &
  76.01 &
  78.67 &
  78.33 &
  79.95 &
  73.81 &
  73.62 &
  76.54 &
  75.84 &
  75.59 &
  74.97 \\
 &
  VGG16 &
  75.11 &
  74.58 &
  73.97 &
  69.34 &
  69.07 &
  71.42 &
  76.67 &
  76.46 &
  75.96 &
  75.12 &
  75.30 &
  75.96 \\ \hline
\multirow{5}{*}{\rotatebox[origin=b]{90}{Minnie}} &
  EfficientNetB0 &
  73.92 &
  73.10 &
  72.03 &
  75.05 &
  75.91 &
  76.83 &
  80.82 &
  79.96 &
  76.49 &
  81.55 &
  78.80 &
  75.41 \\
 &
  MobileNetV2 &
  71.56 &
  71.19 &
  72.38 &
  76.42 &
  71.62 &
  69.63 &
  79.00 &
  78.69 &
  77.29 &
  79.25 &
  78.10 &
  75.96 \\
 &
  MobileNetV3 &
  72.52 &
  64.25 &
  70.62 &
  80.25 &
  75.38 &
  76.69 &
  84.64 &
  71.56 &
  76.89 &
  84.45 &
  68.08 &
  76.67 \\
 &
  ResNet50 &
  77.54 &
  77.73 &
  78.64 &
  80.39 &
  78.53 &
  77.46 &
  75.78 &
  75.37 &
  80.22 &
  78.75 &
  78.83 &
  80.03 \\
 &
  VGG16 &
  80.08 &
  79.67 &
  82.28 &
  79.45 &
  78.70 &
  77.54 &
  83.77 &
  \underline{83.24} &
  76.94 &
  80.03 &
  80.53 &
  80.20 \\ \hline\noalign{\smallskip}
\end{tabular}
\end{table*}

In the second analysis, we focus on the effect of post-training quantization and quantization aware training on robustness. Based on Table~\ref{tab:kth_baseline} we focus on the 'different bot' conditions since changes in viewpoint have a much larger impact on the models' robustness. When analyzing Table~\ref{tab:kth_quant} a similar picture to Table~\ref{tab:quantization_accuracy} can be seen. When only applying post-training quantization (PTQ), huge performance drops can be seen. Especially the MobileNet architectures show several drops in performance. In contrast, EfficientNetB0, VGG16, and ResNet50 do not show major performance drops when only applying PTQ.

Table~\ref{tab:kth_quant} however, shows that post-training quantization (PTQ) and quantization aware training (QAT) can have a significant impact on the robustness of the models. While, in general the quantization aware training does benefit the models' performance, adverse effects can be seen. For the VGG16 models trained with Fine-Tuning/NormAug, the performance drops from 83.24 percent down to 76.91 percent after quantization aware training is performed. Only for MobileNetV3 the quantization aware training always seems to be beneficial.

\renewcommand{\arraystretch}{1.15}
\setlength{\tabcolsep}{2.5pt}
\begin{table*}[ht]
\centering
\caption{Comparsion of the balanced accuracy of the optimizations for different inference accelerators, TRT32: TensorRT FP32, TRT16: TensorRT FP16, Myriad: OpenVino Myriad VPU FP16 on KTH-Idol2 with different bot conditions. All values are in percent.}
\label{tab:kth_optim}
\begin{tabular}{ll|cccc|cccc|cccc|cccc}
 &
   &
  \multicolumn{4}{c|}{L2SP NormAug} &
  \multicolumn{4}{c|}{L2SP ExtraAug} &
  \multicolumn{4}{c|}{Fine-Tuning NormAug} &
  \multicolumn{4}{c}{Fine-Tuning ExtraAug} \\
 &
   &
  \begin{tabular}[c]{@{}c@{}}QAT\\ DBSL\end{tabular} &
  \begin{tabular}[c]{@{}c@{}}TRT32\\ DBSL\end{tabular} &
  \begin{tabular}[c]{@{}c@{}}TRT16\\ DBSL\end{tabular} &
  \begin{tabular}[c]{@{}c@{}}Myriad\\ DBSL\end{tabular} &
  \begin{tabular}[c]{@{}c@{}}QAT\\ DBSL\end{tabular} &
  \begin{tabular}[c]{@{}c@{}}TRT32\\ DBSL\end{tabular} &
  \begin{tabular}[c]{@{}c@{}}TRT16\\ DBSL\end{tabular} &
  \begin{tabular}[c]{@{}c@{}}Myriad\\ DBSL\end{tabular} &
  \begin{tabular}[c]{@{}c@{}}QAT\\ DBSL\end{tabular} &
  \begin{tabular}[c]{@{}c@{}}TRT32\\ DBSL\end{tabular} &
  \begin{tabular}[c]{@{}c@{}}TRT16\\ DBSL\end{tabular} &
  \begin{tabular}[c]{@{}c@{}}Myriad\\ DBSL\end{tabular} &
  \begin{tabular}[c]{@{}c@{}}QAT\\ DBSL\end{tabular} &
  \begin{tabular}[c]{@{}c@{}}TRT32\\ DBSL\end{tabular} &
  \begin{tabular}[c]{@{}c@{}}TRT16\\ DBSL\end{tabular} &
  \begin{tabular}[c]{@{}c@{}}Myriad\\ DBSL\end{tabular} \\ \hline
\multirow{5}{*}{\rotatebox[origin=b]{90}{Dumbo}} &
  EfficientNetB0 &
  \underline{75.16} &
  72.33 &
  72.35 &
  72.80 &
  \underline{75.39} &
  75.08 &
  74.99 &
  75.02 &
  75.66 &
  76.10 &
  76.06 &
  \underline{76.26} &
  74.54 &
  76.51 &
  76.48 &
  \underline{76.92} \\
 &
  MobileNetV2 &
  \underline{71.20} &
  69.21 &
  69.17 &
  67.94 &
  \underline{77.99} &
  73.27 &
  73.43 &
  72.32 &
  71.33 &
  74.03 &
  \underline{74.13} &
  74.06 &
  72.35 &
  75.23 &
  \underline{75.26} &
  74.82 \\
 &
  MobileNetV3 &
  72.39 &
  \underline{74.61} &
  74.57 &
  74.51 &
  74.62 &
  \underline{74.89} &
  74.88 &
  74.70 &
  69.25 &
  76.19 &
  76.19 &
  \underline{76.25} &
  67.87 &
  \underline{78.88} &
  \underline{78.88} &
  78.27 \\
 &
  ResNet50 &
  \underline{79.58} &
  77.23 &
  77.09 &
  77.08 &
  \underline{82.10} &
  81.13 &
  81.11 &
  80.95 &
  \underline{79.16} &
  74.94 &
  74.98 &
  74.83 &
  \underline{80.33} &
  79.02 &
  79.14 &
  79.02 \\
 &
  VGG16 &
  \underline{79.10} &
  77.16 &
  77.12 &
  77.26 &
  \underline{77.22} &
  73.23 &
  73.23 &
  73.24 &
  \underline{80.16} &
  79.95 &
  79.99 &
  79.93 &
  \underline{81.81} &
  78.41 &
  78.33 &
  78.36 \\ \hline
\multirow{5}{*}{\rotatebox[origin=b]{90}{Minnie}} &
  EfficientNetB0 &
  75.91 &
  76.67 &
  76.71 &
  \underline{76.95} &
  \underline{78.15} &
  76.99 &
  77.01 &
  76.92 &
  77.64 &
  \underline{82.39} &
  82.31 &
  82.25 &
  76.87 &
  81.97 &
  81.93 &
  \underline{82.15} \\
 &
  MobileNetV2 &
  76.26 &
  76.50 &
  \underline{76.54} &
  75.75 &
  74.71 &
  79.17 &
  \underline{79.26} &
  78.89 &
  78.24 &
  \underline{81.73} &
  81.65 &
  80.73 &
  78.07 &
  \underline{81.03} &
  80.96 &
  78.75 \\
 &
  MobileNetV3 &
  77.01 &
  \underline{80.16} &
  80.11 &
  80.07 &
  82.22 &
  83.90 &
  \underline{83.98} &
  83.85 &
  80.09 &
  85.48 &
  \underline{85.53} &
  85.04 &
  79.50 &
  \underline{85.18} &
  85.15 &
  85.17 \\
 &
  ResNet50 &
  \underline{82.70} &
  80.71 &
  80.61 &
  80.72 &
  80.96 &
  \underline{83.39} &
  83.35 &
  83.31 &
  \underline{79.46} &
  78.78 &
  78.79 &
  78.84 &
  \underline{81.04} &
  80.17 &
  80.14 &
  80.27 \\
 &
  VGG16 &
  \underline{83.52} &
  80.98 &
  81.01 &
  81.09 &
  77.16 &
  \underline{82.49} &
  82.45 &
  82.45 &
  79.13 &
  84.08 &
  84.04 &
  \underline{84.19} &
  \underline{82.57} &
  81.09 &
  81.05 &
  81.06 \\ \hline\noalign{\smallskip}

 &
   &
  \multicolumn{4}{c|}{L2SP NormAug} &
  \multicolumn{4}{c|}{L2SP ExtraAug} &
  \multicolumn{4}{c|}{Fine-Tuning NormAug} &
  \multicolumn{4}{c}{Fine-Tuning ExtraAug} \\
 &
   &
  \begin{tabular}[c]{@{}c@{}}QAT\\ DBDL\end{tabular} &
  \begin{tabular}[c]{@{}c@{}}TRT32\\ DBDL\end{tabular} &
  \begin{tabular}[c]{@{}c@{}}TRT16\\ DBDL\end{tabular} &
  \begin{tabular}[c]{@{}c@{}}Myriad\\ DBDL\end{tabular} &
  \begin{tabular}[c]{@{}c@{}}QAT\\ DBDL\end{tabular} &
  \begin{tabular}[c]{@{}c@{}}TRT32\\ DBDL\end{tabular} &
  \begin{tabular}[c]{@{}c@{}}TRT16\\ DBDL\end{tabular} &
  \begin{tabular}[c]{@{}c@{}}Myriad\\ DBDL\end{tabular} &
  \begin{tabular}[c]{@{}c@{}}QAT\\ DBDL\end{tabular} &
  \begin{tabular}[c]{@{}c@{}}TRT32\\ DBDL\end{tabular} &
  \begin{tabular}[c]{@{}c@{}}TRT16\\ DBDL\end{tabular} &
  \begin{tabular}[c]{@{}c@{}}Myriad\\ DBDL\end{tabular} &
  \begin{tabular}[c]{@{}c@{}}QAT\\ DBDL\end{tabular} &
  \begin{tabular}[c]{@{}c@{}}TRT32\\ DBDL\end{tabular} &
  \begin{tabular}[c]{@{}c@{}}TRT16\\ DBDL\end{tabular} &
  \begin{tabular}[c]{@{}c@{}}Myriad\\ DBDL\end{tabular} \\ \hline
\multirow{5}{*}{\rotatebox[origin=b]{90}{Dumbo}} &
  EfficientNetB0 &
  \underline{74.12} &
  70.76 &
  70.72 &
  71.46 &
  \underline{73.39} &
  72.95 &
  72.90 &
  73.35 &
  71.46 &
  71.55 &
  71.43 &
  \underline{72.43} &
  72.12 &
  71.78 &
  71.82 &
  \underline{72.81} \\
 &
  MobileNetV2 &
  \underline{72.12} &
  68.99 &
  68.97 &
  68.12 &
  \underline{78.24} &
  73.66 &
  73.69 &
  71.52 &
  70.34 &
  \underline{71.45} &
  71.43 &
  70.18 &
  71.13 &
  73.39 &
  \underline{73.43} &
  72.83 \\
 &
  MobileNetV3 &
  68.84 &
  72.11 &
  72.12 &
  \underline{72.27} &
  71.88 &
  75.33 &
  \underline{75.39} &
  75.04 &
  68.00 &
  \underline{74.27} &
  \underline{74.27} &
  73.79 &
  65.98 &
  75.27 &
  \underline{75.30} &
  74.91 \\
 &
  ResNet50 &
  \underline{76.01} &
  74.08 &
  74.12 &
  74.08 &
  \underline{79.95} &
  78.32 &
  78.34 &
  78.34 &
  \underline{76.54} &
  73.46 &
  73.48 &
  73.48 &
  74.97 &
  75.08 &
  75.04 &
  \underline{75.14} \\
 &
  VGG16 &
  73.97 &
  74.95 &
  74.86 &
  \underline{75.02} &
  \underline{71.42} &
  69.21 &
  69.21 &
  69.21 &
  75.96 &
  \underline{76.39} &
  76.29 &
  76.22 &
  \underline{75.96} &
  75.06 &
  75.07 &
  75.17 \\ \hline
\multirow{5}{*}{\rotatebox[origin=b]{90}{Minnie}} &
  EfficientNetB0 &
  72.03 &
  74.48 &
  74.52 &
  \underline{74.80} &
  \underline{76.83} &
  74.73 &
  74.82 &
  75.51 &
  76.49 &
  80.53 &
  80.59 &
  \underline{80.62} &
  75.41 &
  81.78 &
  81.75 &
  \underline{81.95} \\
 &
  MobileNetV2 &
  \underline{72.38} &
  71.10 &
  71.22 &
  69.60 &
  69.63 &
  76.84 &
  \underline{76.88} &
  75.73 &
  77.29 &
  \underline{79.17} &
  79.15 &
  77.19 &
  75.96 &
  \underline{79.24} &
  79.17 &
  77.43 \\
 &
  MobileNetV3 &
  70.62 &
  \underline{72.07} &
  71.90 &
  72.00 &
  76.69 &
  80.29 &
  \underline{80.33} &
  80.26 &
  76.89 &
  83.83 &
  \underline{83.85} &
  83.30 &
  76.67 &
  \underline{83.93} &
  83.65 &
  83.74 \\
 &
  ResNet50 &
  \underline{78.64} &
  76.83 &
  76.89 &
  76.87 &
  77.46 &
  79.50 &
  79.50 &
  \underline{79.56} &
  \underline{80.22} &
  75.85 &
  75.95 &
  75.88 &
  \underline{80.03} &
  78.84 &
  78.79 &
  78.75 \\
 &
  VGG16 &
  \underline{82.28} &
  80.69 &
  80.71 &
  80.72 &
  77.54 &
  \underline{80.87} &
  \underline{80.87} &
  \underline{80.87} &
  76.94 &
  83.42 &
  83.33 &
  \underline{83.54} &
  80.20 &
  81.56 &
  81.61 &
  \underline{81.64} \\ \hline\noalign{\smallskip}
\end{tabular}
\end{table*}

When comparing the impact of the different inference optimization techniques in Table~\ref{tab:kth_optim} a stronger impact of the 8-bit quantization can be observed. In contrast to the baseline optimization comparison in Table~\ref{tab:optim_accuracy} more significant performance deviations can be seen when comparing quantization aware training results with TRT32, TRT16, and Myriad optimization, which still use Float to store weights and activations. Especially the mobile-ready architectures MobileNetV2/V3 and EfficientNetB0 show up to 10 percent improved balanced accuracy when encountered with viewpoint and lighting changes.

When comparing the sensitivity of the different architectures, a similar effect to the KTH baseline accuracy can be observed (see Table~\ref{tab:kth_baseline}). In particular, the MobileNetV3 models show good robustness against viewpoint and lighting changes. It also can be seen that the Fine-Tuning / ExtraAug strategy, when applied to the mobile architectures, benefits from the floating-point optimization.

Furthermore, it can be seen that FP32 and FP16 both work equally well in terms of robustness. Little to no difference can be seen between TRT FP32, TRT FP16, and the Myriad FP16 models. Only for the MobileNetV2 models, a small difference between the TRT FP16 and Myriad FP16 models can be observed.

\subsection{Inference evaluation}
In our last analysis, we focus on the inference performance of the models using different optimization techniques. Since the inference performance is crucial on a model robot platform, we performed all tests using either the Raspberry Pi with attached USB accelerators or the Nvidia Jetson Nano platform. We repeated the inference test using different models to ensure that our different transfer learning strategies did not negatively influence the inference performance. No differences could be seen when comparing the different transfer learning strategies, and all models performed equally regardless of the transfer learning strategy. All CPU models were fully quantized 8-bit TensorFlow Lite models.

As seen in Table~\ref{tab:inference_performance} there is a very large difference between mobile-ready CNN architectures and our baseline ResNet50 and VGG16 architectures. On average, the ResNet50 needed 12.17 seconds to process a batch of 32 images when executed only on CPU, while the VGG16 model needed 38.63 seconds. In contrast, the MobileNetv3 only needed between 1.97 to 2.53 seconds to process 32 images even when executed on CPU only. From the comparison, it becomes clear that mobile-ready networks are necessary when aiming for a real-time application in mobile robotics.

When comparing the performance of the models on the inference accelerators, a significant performance boost over the CPU can be seen as expected. However, even when using an inference accelerator, the ResNet50 and VGG16 are very slow compared to mobile-ready networks. Only the Jetson Nano with FP16 and the ResNet50 achieved a real-time ready throughput of 25 frames per second.

It can also be seen that the number of output classes only has little to no impact on the models' inference performance. When comparing the dataset most (MIT67) and the least classes (KTH-Idol2), no significant differences in processing time and theoretical throughput can be observed. The additional calculations needed for the 67 output classes are relatively small in comparison to the overall FLOPS.

When comparing the different optimization techniques, it can be seen that the Coral TPU stick and the Jetson Nano models are very close together. On the Jetson Nano for the mobile-ready architectures, the difference between FP32 and FP16 is relatively small. However, for the large architectures, a clear benefit when using FP16 can be seen. For example, with the VGG16 architecture Event8 dataset, a 68 percent increase in throughput can be seen when using FP16 instead of FP32. Similar increases can also be seen for the ResNet50 models.

\renewcommand{\arraystretch}{1.15}
\begin{table*}[ht]
\centering
\caption{Inference performance of the different models on mobile platforms. Time Elapsed denoted the average time to process a batch of 32 images (lower is better). Throughput denotes the theoretical throughput of images per second (higher is better). All values are average across 10 runs.}
\label{tab:inference_performance}
\begin{tabular}{ll|cc|cc|cc|cc|cc}
 &
   &
  \multicolumn{2}{c|}{Raspberry Pi CPU} &
  \multicolumn{2}{c|}{Myriad VPU} &
  \multicolumn{2}{c|}{Coral TPU Stick} &
  \multicolumn{2}{c|}{Jetson Nano FP32} &
  \multicolumn{2}{c}{Jetson Nano FP16} \\
 &
   &
  \begin{tabular}[c]{@{}c@{}}Time\\ Elapsed\end{tabular} &
  \begin{tabular}[c]{@{}c@{}}Through-\\ put\end{tabular} &
  \begin{tabular}[c]{@{}c@{}}Time\\ Elapsed\end{tabular} &
  \begin{tabular}[c]{@{}c@{}}Through-\\ put\end{tabular} &
  \begin{tabular}[c]{@{}c@{}}Time\\ Elapsed\end{tabular} &
  \begin{tabular}[c]{@{}c@{}}Through-\\ put\end{tabular} &
  \begin{tabular}[c]{@{}c@{}}Time\\ Elapsed\end{tabular} &
  \begin{tabular}[c]{@{}c@{}}Through-\\ put\end{tabular} &
  \begin{tabular}[c]{@{}c@{}}Time\\ Elapsed\end{tabular} &
  \begin{tabular}[c]{@{}c@{}}Through-\\ put\end{tabular} \\ \hline
\multirow{5}{*}{\rotatebox[origin=b]{90}{Event8}}     & EfficientNetB0 & 2.51  & 12.76 & 0.90 & 35.63 & 0.64 & 50.06 & 0.58 & 49.95 & 0.56 & 51.81 \\
                            & MobileNetV2    & 2.43  & 13.17 & 1.09 & 29.49 & 0.92 & 34.61 & 0.79 & 36.78 & 0.76 & 38.31 \\
                            & MobileNetV3    & 1.98  & 16.19 & 0.73 & 43.74 & 0.60 & 52.95 & 0.50 & 58.53 & 0.46 & 63.93 \\
                            & ResNet50       & 39.01 & 0.82  & 4.83 & 6.62  & 2.00 & 16.02 & 3.98 & 7.31  & 2.40 & 12.13 \\
                            & VGG16          & 12.20 & 2.62  & 2.01 & 15.95 & 2.66 & 12.02 & 1.99 & 14.65 & 1.13 & 25.85 \\ \hline
\multirow{5}{*}{\rotatebox[origin=b]{90}{Scene15}}    & EfficientNetB0 & 2.53  & 12.65 & 0.89 & 35.86 & 0.59 & 54.11 & 0.59 & 49.71 & 0.56 & 51.50 \\
                            & MobileNetV2    & 2.44  & 13.09 & 1.09 & 29.48 & 0.93 & 34.52 & 0.78 & 37.20 & 0.74 & 39.11 \\
                            & MobileNetV3    & 1.99  & 16.10 & 0.73 & 43.55 & 0.64 & 50.37 & 0.49 & 58.99 & 0.46 & 62.95 \\
                            & ResNet50       & 38.73 & 0.83  & 4.85 & 6.60  & 2.02 & 15.82 & 4.02 & 7.24  & 2.44 & 11.92 \\
                            & VGG16          & 12.06 & 2.65  & 2.01 & 15.94 & 2.66 & 12.02 & 1.99 & 14.58 & 1.13 & 25.79 \\ \hline
\multirow{5}{*}{\rotatebox[origin=b]{90}{Stanford40}} & EfficientNetB0 & 2.52  & 12.71 & 0.90 & 35.72 & 0.61 & 52.23 & 0.57 & 50.72 & 0.56 & 52.09 \\
                            & MobileNetV2    & 2.44  & 13.09 & 1.09 & 29.37 & 0.92 & 34.72 & 0.78 & 37.08 & 0.76 & 38.43 \\
                            & MobileNetV3    & 1.99  & 16.10 & 0.74 & 43.31 & 0.63 & 51.15 & 0.50 & 57.68 & 0.45 & 64.29 \\
                            & ResNet50       & 38.30 & 0.84  & 4.84 & 6.61  & 2.02 & 15.81 & 4.02 & 7.25  & 2.44 & 11.92 \\
                            & VGG16          & 12.00 & 2.67  & 2.01 & 15.94 & 2.67 & 12.00 & 1.98 & 14.71 & 1.12 & 25.98 \\ \hline
\multirow{5}{*}{\rotatebox[origin=b]{90}{MIT67}}      & EfficientNetB0 & 2.51  & 12.76 & 0.90 & 35.70 & 0.64 & 50.00 & 0.59 & 49.65 & 0.57 & 50.89 \\
                            & MobileNetV2    & 2.42  & 13.21 & 1.09 & 29.46 & 0.93 & 34.49 & 0.77 & 37.56 & 0.76 & 38.12 \\
                            & MobileNetV3    & 1.97  & 16.26 & 0.73 & 43.61 & 0.62 & 51.90 & 0.50 & 57.79 & 0.45 & 63.95 \\
                            & ResNet50       & 38.41 & 0.83  & 4.85 & 6.60  & 2.02 & 15.81 & 4.00 & 7.27  & 2.44 & 11.90 \\
                            & VGG16          & 12.05 & 2.66  & 2.00 & 15.97 & 2.67 & 11.99 & 1.99 & 14.62 & 1.13 & 25.74 \\ \hline
\multirow{5}{*}{\rotatebox[origin=b]{90}{KTH-Idol2}}   & EfficientNetB0 & 2.51  & 12.76 & 0.90 & 35.69 & 0.64 & 49.74 & 0.59 & 49.25 & 0.56 & 52.25 \\
                            & MobileNetV2    & 2.44  & 13.14 & 1.08 & 29.51 & 0.92 & 34.84 & 0.79 & 36.77 & 0.73 & 39.80 \\
                            & MobileNetV3    & 1.98  & 16.16 & 0.73 & 43.55 & 0.63 & 50.69 & 0.54 & 53.77 & 0.45 & 64.61 \\
                            & ResNet50       & 38.38 & 0.83  & 4.75 & 6.73  & 1.95 & 16.44 & 4.04 & 7.21  & 2.42 & 12.00 \\
                            & VGG16          & 11.93 & 2.68  & 2.00 & 15.98 & 2.67 & 12.00 & 1.97 & 14.74 & 1.13 & 25.82 \\ \hline\noalign{\smallskip}
\end{tabular}
\end{table*}

\section{Discussion}
The evaluation of our four transfer learning strategies shows that despite the risk of the loss of co-adaption between filters, Fine-Tuning outperforms the explicit inductive bias. Furthermore, it has been shown that the models benefit from a more aggressive data augmentation, especially when the number of classes rises. The statistical evaluation has shown both small effect sizes for the ExtraAug and the Fine-Tuning strategy. The slopegraph in Fig~\ref{fig:dabest_strategies} and Table~\ref{tab:baseline_accuracy} show that the ExtraAug augmentation policy can increase the balanced accuracy in the MIT67 dataset up to 79.19 percent, which is a state-of-the-art results for a single patch place recognition network \cite{seong2020a}. Also, when looking at Table~\ref{tab:kth_baseline} and Fig.~\ref{fig:dabest_kth} it can be seen that Fine-Tuning and ExtraAug are beneficial for the models' robustness for the mobile-ready architectures.

When using ExtraAug, a quantization aware training is very beneficial for all tested CNN architectures, especially for the EfficientNetB0 architecture. Without the quantization aware training, high drops in accuracy are possible and likely for datasets with a high number of classes when quantizing the networks. When only looking at the generalization performance evaluation, the quantization aware training almost negated the negative effect of 8-bit quantization. Only in rare cases, do the FP32 and FP16 outperform the 8-bit quantized models. However, the robustness evaluation clearly showed that quantization aware training is good for the generalization performance but is not beneficial for the robustness (see Table~\ref{tab:kth_quant}). Especially when using Fine-Tuning the quantization aware training can even worsen the performance compared to simple post-training quantization. Furthermore, the results show that the balanced accuracy of the mobile-ready architectures can largely drop when using 8-bit quantization. However, this effect can be ignored using optimization, which uses floating-point values to store weights and activations. In particular, the MobileNet architectures benefit from floating-point-based optimizations like TensorRT.

While being heavy on computational resources, the VGG16 models show good robustness against viewpoint and lighting changes even after quantization. Also, MobileNetV3 shows good robustness, comparable to the much larger ResNet50 architecture. EfficientNetB0 still remains a good choice but with lesser robustness. If only CPU is available, an 8-bit quantized EfficientNetB0 with quantization aware training is a good mix of accuracy and throughput. If a neural network accelerator is available, MobileNetV3 models tend to outperform the EfficientNetB0 in terms of robustness and throughput.

The inference performance results show that the number of classes as output does not significantly impact the inference performance. The impact of the base architecture, however, is significantly larger. The results further showed that neither the VGG16 nor the ResNet50 are feasible for mobile use. A theoretical throughput of 25.98 frames per second using VGG16 could be achieved on a Jetson Nano 4GB with an optimized TensorRT FP16 model. While this would be enough for real-time processing of a standard video capture, it must be noted that these performance values have been achieved with preprocessed images loaded directly from RAM and without other background processes. Therefore the achievable throughput in a real-life mobile robot scenario is likely to be lower. Using ResNet50, none of the optimization techniques reached 25 frames per second or higher. Even inference accelerators cannot speed up large-scale architectures like ResNet50 and VGG16 to a real-time ready level. Also, this currently only includes place recognition. For processes where multiple models are necessary large-scale architectures are most likely not feasible. If only CPU is available, an 8-bit quantized EfficientNetB0 with quantization aware training is a good mix of accuracy and throughput, while all mobile-ready architectures achieve good throughput when an inference accelerator is available. While FP16 on Jetson Nano speeds up the throughput for all models, this speed-up is significantly large only for large-scale architectures.

\section{Conclusion}
We demonstrate that using Fine-Tuning in combination with heavy data augmentation, state-of-the-art results for visual place recognition can be achieved in various datasets, using widely available mobile-ready CNN architectures. By combining powerful architectures with our transfer learning strategy, we created models showing state-of-the-art accuracy for single patch models while still showing good inference performance on real mobile platforms.

We showed that more modern mobile-ready models tend to outperform the traditional baseline architectures VGG16 and ResNet50 in all our tested scenarios, both in terms of accuracy and speed. While more powerful mobile-ready architectures are available, VGG16 and ResNet50 still are common choices when performing transfer learning. Due to their ease of use and relatively simple architectures, they are easy to train and optimize for inference. Even inference accelerators cannot speed up large-scale architectures like ResNet50 and VGG16 to a real-time ready level.

However, other powerful and modern mobile ready achieve very good results both in accuracy and inference performance. As shown in the generalization performance evaluation, the Fine-Tuning / ExtraAug transfer learning strategy combined with the EfficientNetB0 architecture outperforms most other architectures and strategies. When applying transfer learning, it is a solid choice, shows good accuracy with good robustness, and is very fast when using an inference accelerator. If only CPU is available, an 8-bit quantized EfficientNetB0 with quantization aware training is a good mix of accuracy and throughput. If a neural network accelerator is available, MobileNetV3 models tend to outperform the EfficientNetB0 in terms of robustness and throughput.

\subsection{Limitations}
The tested CNN models are currently all pre-trained on the ImageNet dataset. However, other studies show that a specialized pre-training, for example, on the Places365 dataset, can be beneficial \cite{mancini2017a, zhou2018a}. Furthermore, the performance is currently tested on the place recognition performance. Yet the effectiveness of our transfer learning strategies on other common tasks like object detection or image segmentation has not yet been tested.

In addition, our comparison of the inference performance is based upon entry-level hardware. However, other more powerful inference accelerators such as the Nvidia Jetson TX2 or the Google Coral Dev Board are available. Therefore, real-time inference with large-scale architectures could also be achieved by using a more powerful but more expensive hardware platform.

\subsection{Future work}
In order to fully assess the generalization capabilities of our transfer learning strategies, we will re-evaluate the performance using networks pre-trained with the Place365 dataset \cite{zhou2018a}. Additional other vision-based tasks such as object detection or image segmentation are common in combination with place recognition. Other work shows that our tested mobile CNN architectures are also a suitable base for vision tasks like object detection \cite{sandler2018a, mingxing2019a, howard2019a}. Therefore in future research, we will examine the effectiveness of our transfer learning strategies for combined tasks such as simultaneous place recognition and object detection on mobile platforms.

\appendices
\section*{Acknowledgment}
This research is partly funded by the German Federal Ministry of Education and Research (no. 13FH176PX8, no. 13FH4I05IA, no. 13FH566KX9).

\ifCLASSOPTIONcaptionsoff
  \newpage
\fi

\bibliographystyle{IEEEtran}
\bibliography{Literat,IEEEabrv}

\begin{thebibliography}{10}
\providecommand{\url}[1]{#1}
\csname url@samestyle\endcsname
\providecommand{\newblock}{\relax}
\providecommand{\bibinfo}[2]{#2}
\providecommand{\BIBentrySTDinterwordspacing}{\spaceskip=0pt\relax}
\providecommand{\BIBentryALTinterwordstretchfactor}{4}
\providecommand{\BIBentryALTinterwordspacing}{\spaceskip=\fontdimen2\font plus
\BIBentryALTinterwordstretchfactor\fontdimen3\font minus
  \fontdimen4\font\relax}
\providecommand{\BIBforeignlanguage}[2]{{%
\expandafter\ifx\csname l@#1\endcsname\relax
\typeout{** WARNING: IEEEtran.bst: No hyphenation pattern has been}%
\typeout{** loaded for the language `#1'. Using the pattern for}%
\typeout{** the default language instead.}%
\else
\language=\csname l@#1\endcsname
\fi
#2}}
\providecommand{\BIBdecl}{\relax}
\BIBdecl

\bibitem{qin2018a}
T.~Qin, P.~Li, and S.~Shen, ``{VINS-Mono: A Robust and Versatile Monocular
  Visual-Inertial State Estimator},'' \emph{{IEEE Trans. Robot.}}, vol.~34,
  no.~4, pp. 1004--1020, 2018.

\bibitem{zhao2020a}
Y.~Zhao and P.~A. Vela, ``{Good Feature Matching: Toward Accurate, Robust
  {VO}/{VSLAM} With Low Latency},'' \emph{{IEEE Trans. Robot.}}, vol.~36,
  no.~3, pp. 657--675, 2020.

\bibitem{wen2020a}
F.~Wen, R.~Ying, Z.~Gong, and P.~Liu, ``{Efficient Algorithms for Maximum
  Consensus Robust Fitting},'' \emph{{IEEE Trans. Robot.}}, vol.~36, no.~1, pp.
  92--106, 2020.

\bibitem{cadena2016a}
C.~Cadena, L.~Carlone, H.~Carrillo, Y.~Latif, D.~Scaramuzza, J.~Neira, I.~Reid,
  and J.~J. Leonard, ``{Past, Present, and Future of Simultaneous Localization
  and Mapping: Toward the Robust-Perception Age},'' \emph{{IEEE Trans.
  Robot.}}, vol.~32, no.~6, pp. 1309--1332, 2016.

\bibitem{bescos2021a}
B.~Bescos, C.~Cadena, and J.~Neira, ``{Empty Cities: A Dynamic-Object-Invariant
  Space for Visual {SLAM}},'' \emph{{IEEE Trans. Robot.}}, vol.~37, no.~2, pp.
  433--451, 2021.

\bibitem{khaliq2020a}
A.~Khaliq, S.~Ehsan, Z.~Chen, M.~Milford, and K.~McDonald-Maier, ``{A Holistic
  Visual Place Recognition Approach Using Lightweight {CNNs} for Significant
  {ViewPoint} and Appearance Changes},'' \emph{{IEEE Trans. Robot.}}, vol.~36,
  no.~2, pp. 561--569, 2020.

\bibitem{lowry2016a}
S.~Lowry, N.~Sunderhauf, P.~Newman, J.~J. Leonard, D.~Cox, P.~Corke, and M.~J.
  Milford, ``{Visual Place Recognition: A Survey},'' \emph{{IEEE Trans.
  Robot.}}, vol.~32, no.~1, pp. 1--19, 2016.

\bibitem{suenderhauf2015a}
N.~Suenderhauf, S.~Shirazi, F.~Dayoub, B.~Upcroft, and M.~Milford, ``{On the
  performance of {ConvNet} features for place recognition},'' in \emph{IROS
  2015 Proc.}\hskip 1em plus 0.5em minus 0.4em\relax {IEEE}, 2015, pp.
  4297--4304.

\bibitem{arroyo2016a}
R.~Arroyo, P.~F. Alcantarilla, L.~M. Bergasa, and E.~Romera, ``{Fusion and
  Binarization of {CNN} Features for Robust Topological Localization across
  Seasons},'' in \emph{IROS 2016 Proc.}\hskip 1em plus 0.5em minus 0.4em\relax
  {IEEE}, 2016, pp. 4656--4663.

\bibitem{mancini2017a}
M.~Mancini, S.~R. Bulo, E.~Ricci, and B.~Caputo, ``{Learning Deep {NBNN}
  Representations for Robust Place Categorization},'' \emph{{IEEE} Robot.
  Autom. Lett.}, vol.~2, no.~3, pp. 1794--1801, 2017.

\bibitem{xie2020a}
L.~Xie, F.~Lee, L.~Liu, K.~Kotani, and Q.~Chen, ``{Scene recognition: A
  comprehensive survey},'' \emph{{Pattern Recognit.}}, vol. 102, p. 107205,
  2020.

\bibitem{oore1997a}
S.~Oore, G.~E. Hinton, and G.~Dudek, ``{A Mobile Robot That Learns Its
  Place},'' \emph{{Neural Comput.}}, vol.~9, no.~3, pp. 683--699, 1997.

\bibitem{sanchez2012a}
A.~Sanchez, A.~d. Castro, S.~Elvira, G.~Glez-de Rivera, and J.~Garrido,
  ``{Autonomous indoor ultrasonic positioning system based on a low-cost
  conditioning circuit},'' \emph{Measurement}, vol.~45, no.~3, pp. 276--283,
  2012.

\bibitem{zhangji2015a}
J.~Zhang and S.~Singh, ``{Visual-lidar Odometry and Mapping: Low-drift, Robust,
  and Fast},'' in \emph{ICRA 2015 Proc.}\hskip 1em plus 0.5em minus 0.4em\relax
  {IEEE}, 2015, pp. 2174--2181.

\bibitem{agrawal2006a}
M.~Agrawal and K.~Konolige, ``Real-time localization in outdoor environments
  using stereo vision and inexpensive {GPS},'' in \emph{ICPR '06 Proc.}\hskip
  1em plus 0.5em minus 0.4em\relax {IEEE}, 2006, pp. 1063--1068.

\bibitem{leCun2015a}
Y.~LeCun, Y.~Bengio, and G.~Hinton, ``Deep learning,'' \emph{Nature}, vol. 521,
  no. 7553, pp. 436--444, 2015.

\bibitem{pan2010a}
S.~J. Pan and Q.~Yang, ``{A Survey on Transfer Learning},'' \emph{{IEEE} T.
  Knowl. Data. En.}, vol.~22, no.~10, pp. 1345--1359, 2010.

\bibitem{li2018a}
X.~Li, Y.~Grandvalet, and F.~Davoine, ``{Explicit Inductive Bias for Transfer
  Learning with Convolutional Networks},'' in \emph{PMLR 2018 Proc.}, 2018, pp.
  2825--2834.

\bibitem{seong2020a}
H.~Seong, J.~Hyun, and E.~Kim, ``{FOSNet: An End-to-End Trainable Deep Neural
  Network for Scene Recognition},'' \emph{{IEEE Access}}, vol.~8, pp.
  82\,066--82\,077, 2020.

\bibitem{sandler2018a}
M.~Sandler, A.~Howard, M.~Zhu, A.~Zhmoginov, and L.-C. Chen, ``{MobileNetV}2:
  Inverted residuals and linear bottlenecks,'' in \emph{CVPR 2018 Proc.}\hskip
  1em plus 0.5em minus 0.4em\relax {IEEE}, 2018, pp. 4510--4520.

\bibitem{howard2019a}
A.~Howard, M.~Sandler, B.~Chen, W.~Wang, L.-C. Chen, M.~Tan, G.~Chu,
  V.~Vasudevan, Y.~Zhu, R.~Pang, H.~Adam, and Q.~Le, ``{Searching for
  MobileNetV3},'' in \emph{ICCV 2019 Proc.}\hskip 1em plus 0.5em minus
  0.4em\relax {IEEE}, 2019, pp. 1314--1324.

\bibitem{mingxing2019a}
T.~Mingxing and L.~Quoc, ``{EfficientNet: Rethinking Model Scaling for
  Convolutional Neural Networks},'' in \emph{ICML 2019 Proc.}, vol.~97.\hskip
  1em plus 0.5em minus 0.4em\relax {PMLR}, 2019, pp. 6105--6114.

\bibitem{li2007a}
L.-J. Li and L.~Fei-Fei, ``{What, where and who? Classifying events by scene
  and object recognition},'' in \emph{ICCV 2007 Proc.}\hskip 1em plus 0.5em
  minus 0.4em\relax {IEEE}, 2007, pp. 1--8.

\bibitem{lazebnik2006a}
S.~Lazebnik, C.~Schmid, and J.~Ponce, ``{Beyond Bags of Features: Spatial
  Pyramid Matching for Recognizing Natural Scene Categories},'' in \emph{CVPR
  2006 Proc.}\hskip 1em plus 0.5em minus 0.4em\relax {IEEE}, 2006, pp.
  2169--2178.

\bibitem{yao2011a}
B.~Yao, X.~Jiang, A.~Khosla, A.~L. Lin, L.~Guibas, and L.~Fei-Fei, ``Human
  action recognition by learning bases of action attributes and parts,'' in
  \emph{ICCV 2011 Proc.}\hskip 1em plus 0.5em minus 0.4em\relax {IEEE}, 2011,
  pp. 1331--1338.

\bibitem{quattoni2009a}
A.~Quattoni and A.~Torralba, ``{Recognizing Indoor Scenes},'' in \emph{CVPR
  2009 Proc.}\hskip 1em plus 0.5em minus 0.4em\relax IEEE, 2009, pp. 413--421.

\bibitem{luo2007a}
J.~Luo, A.~Pronobis, B.~Caputo, and P.~Jensfelt, ``{Incremental learning for
  place recognition in dynamic environments},'' in \emph{{IEEE}/{RSJ} IROS 2007
  Proc.}, 2007, pp. 721--728.

\bibitem{khan2016a}
S.~H. Khan, M.~Hayat, M.~Bennamoun, R.~Togneri, and F.~A. Sohel, ``{A
  Discriminative Representation of Convolutional Features for Indoor Scene
  Recognition},'' \emph{{IEEE Trans. Image Process.}}, vol.~25, no.~7, pp.
  3372--3383, 2016.

\bibitem{xie2020b}
L.~Xie, F.~Lee, L.~Liu, Z.~Yin, and Q.~Chen, ``{Hierarchical Coding of
  Convolutional Features for Scene Recognition},'' \emph{{IEEE Trans.
  Multimedia}}, vol.~22, no.~5, pp. 1182--1192, 2020.

\bibitem{shi2019a}
J.~Shi, H.~Zhu, S.~Yu, W.~Wu, and H.~Shi, ``{Scene Categorization Model Using
  Deep Visually Sensitive Features},'' \emph{{IEEE Access}}, vol.~7, pp.
  45\,230--45\,239, 2019.

\bibitem{cheng2018b}
X.~Cheng, J.~Lu, J.~Feng, B.~Yuan, and J.~Zhou, ``{Scene recognition with
  objectness},'' \emph{{Pattern Recognit.}}, vol.~74, pp. 474--487, 2018.

\bibitem{herranz2016a}
L.~Herranz, S.~Jiang, and X.~Li, ``{Scene Recognition with {CNNs}: Objects,
  Scales and Dataset Bias},'' in \emph{{CVPR 2016 Proc.}}\hskip 1em plus 0.5em
  minus 0.4em\relax {IEEE}, 2016, pp. 571--579.

\bibitem{ryu2018a}
J.~Ryu, M.-H. Yang, and J.~Lim, ``{DFT-based Transformation Invariant Pooling
  Layer for Visual Classification},'' in \emph{{ECCV 2018 Proc.}}\hskip 1em
  plus 0.5em minus 0.4em\relax Springer, 2018, pp. 89--104.

\bibitem{jiang2019a}
S.~Jiang, G.~Chen, X.~Song, and L.~Liu, ``{Deep Patch Representations with
  Shared Codebook for Scene Classification},'' \emph{{ACM Trans. Multimedia
  Comput. Commun. Appl.}}, vol.~15, no.~1s, pp. 1--17, 2019.

\bibitem{he2016a}
K.~He, X.~Zhang, S.~Ren, and J.~Sun, ``Deep residual learning for image
  recognition,'' in \emph{CVPR 2016 Proc.}, 2016, pp. 770--778.

\bibitem{xie2017a}
S.~Xie, R.~Girshick, P.~Dollar, Z.~Tu, and K.~He, ``{Aggregated Residual
  Transformations for Deep Neural Networks},'' in \emph{CVPR 2017 Proc.}\hskip
  1em plus 0.5em minus 0.4em\relax {IEEE}, 2017, pp. 5987--5995.

\bibitem{chollet2017a}
F.~Chollet, ``Xception: Deep learning with depthwise separable convolutions,''
  in \emph{CVPR 2017 Proc.}, 2017, pp. 1800--1807.

\bibitem{szegedy2015a}
C.~Szegedy, W.~Liu, Y.~Jia, P.~Sermanet, S.~Reed, D.~Anguelov, D.~Erhan,
  V.~Vanhoucke, and A.~Rabinovich, ``Going deeper with convolutions,'' in
  \emph{CVPR 2015 Proc.}, 2015, pp. 1--9.

\bibitem{fazl_ersi2012a}
E.~Fazl-Ersi and J.~K. Tsotsos, ``{Histogram of Oriented Uniform Patterns for
  robust place recognition and categorization},'' \emph{Int. J. Robot. Res.},
  vol.~31, no.~4, pp. 468--483, 2012.

\bibitem{oliva2001a}
A.~Oliva and A.~Torralba, ``{Modeling the Shape of the Scene: A Holistic
  Representation of the Spatial Envelope},'' \emph{Int. J. Comput. Vis.},
  vol.~42, no.~3, pp. 145--175, 2001.

\bibitem{lowe2004a}
D.~G. Lowe, ``{Distinctive Image Features from Scale-Invariant Keypoints},''
  \emph{Int. J. Comput. Vis.}, vol.~60, no.~2, pp. 91--110, 2004.

\bibitem{bay2006a}
H.~Bay, T.~Tuytelaars, and L.~Van~Gool, ``{SURF}: Speeded up robust features,''
  in \emph{ECCV '06 Proc.}\hskip 1em plus 0.5em minus 0.4em\relax Springer,
  2006, pp. 404--417.

\bibitem{wu2011a}
J.~Wu and J.~M. Rehg, ``{CENTRIST: A Visual Descriptor for Scene
  Categorization.}'' \emph{IEEE Trans. Pattern Anal. Mach. Intell.}, vol.~33,
  pp. 1489--1501, 2011.

\bibitem{rublee2011a}
E.~Rublee, V.~Rabaud, K.~Konolige, and G.~Bradski, ``{ORB}: An efficient
  alternative to {SIFT} or {SURF},'' in \emph{ICCV 2011 Proc.}\hskip 1em plus
  0.5em minus 0.4em\relax {IEEE}, 2011, pp. 2564--2571.

\bibitem{yosinski2014a}
J.~Yosinski, J.~Clune, Y.~Bengio, and H.~Lipson, ``{How Transferable Are
  Features in Deep Neural Networks?}'' in \emph{NIPS '14 Proc.}\hskip 1em plus
  0.5em minus 0.4em\relax MIT, 2014, pp. 3320--3328.

\bibitem{weiss2016a}
K.~Weiss, T.~M. Khoshgoftaar, and D.~Wang, ``{A survey of transfer learning},''
  \emph{J. Big Data}, vol.~3, no.~9, 2016.

\bibitem{donahue2014a}
J.~Donahue, Y.~Jia, O.~Vinyals, J.~Hoffman, N.~Zhang, E.~Tzeng, and T.~Darrell,
  ``{DeCAF: A Deep Convolutional Activation Feature for Generic Visual
  Recognition},'' in \emph{ICML 2014 Proc.}\hskip 1em plus 0.5em minus
  0.4em\relax JMLR, 2014, pp. I--647--I--655.

\bibitem{wozniak2018a}
P.~Wozniak, H.~Afrisal, R.~G. Esparza, and B.~Kwolek, ``Scene recognition for
  indoor localization of mobile robots using deep {CNN},'' in \emph{ICCVG 2018
  Proc.}\hskip 1em plus 0.5em minus 0.4em\relax Springer, 2018, pp. 137--147.

\bibitem{simonyan2015a}
K.~Simonyan and A.~Zisserman, ``{Very Deep Convolutional Networks for
  Large-Scale Image Recognition},'' in \emph{ICLR 2015 Proc.}, 2015.

\bibitem{cubuk2020a}
E.~D. Cubuk, B.~Zoph, J.~Shlens, and Q.~Le, ``{RandAugment: Practical Automated
  Data Augmentation with a Reduced Search Space},'' in \emph{NIPS 2020 Proc.},
  vol.~33.\hskip 1em plus 0.5em minus 0.4em\relax Curran Associates, Inc.,
  2020, pp. 18\,613--18\,624.

\bibitem{krizhevsky2012a}
A.~Krizhevsky, I.~Sutskever, and G.~E. Hinton, ``{ImageNet Classification with
  Deep Convolutional Neural Networks},'' in \emph{NIPS '12 Proc.}, 2012, pp.
  1097--1105.

\bibitem{buslaev2020a}
A.~Buslaev, V.~I. Iglovikov, E.~Khvedchenya, A.~Parinov, M.~Druzhinin, and
  A.~A. Kalinin, ``{Albumentations: Fast and Flexible Image Augmentations},''
  \emph{Information}, vol.~11, no.~2, p. 125, 2020.

\bibitem{howard2017a}
A.~G. Howard, M.~Zhu, B.~Chen, D.~Kalenichenko, W.~Wang, T.~Weyand,
  M.~Andreetto, and H.~Adam, ``{MobileNets: Efficient Convolutional Neural
  Networks for Mobile Vision Applications},'' \emph{arXiv e-prints}, p.
  arXiv:1704.04861, 2017.

\bibitem{zhou2018a}
B.~Zhou, A.~Lapedriza, A.~Khosla, A.~Oliva, and A.~Torralba, ``{Places: A 10
  Million Image Database for Scene Recognition},'' \emph{IEEE Trans. Pattern
  Anal. Mach. Intell.}, vol.~40, no.~6, pp. 1452--1464, 2018.

\bibitem{arnold2017a}
S.~Arnold and K.~Yamazaki, ``Real-time scene parsing by means of a
  convolutional neural network for mobile robots in disaster scenarios,'' in
  \emph{{ICIA} 2017 Proc.}, 2017, pp. 201--207.

\bibitem{ba2014a}
L.~J. Ba and R.~Caruana, ``{Do Deep Nets Really Need to Be Deep?}'' in
  \emph{NIPS '14 Proc.}\hskip 1em plus 0.5em minus 0.4em\relax MIT, 2014, pp.
  2654--2662.

\bibitem{cheng2018a}
J.~Cheng, J.~Wu, C.~Leng, Y.~Wang, and Q.~Hu, ``{Quantized {CNN}: A Unified
  Approach to Accelerate and Compress Convolutional Networks},'' \emph{{IEEE}
  Trans. Neural Netw. Learn. System.}, vol.~29, no.~10, pp. 4730--4743, 2018.

\bibitem{han2016b}
S.~Han, X.~Liu, H.~Mao, J.~Pu, A.~Pedram, M.~A. Horowitz, and W.~J. Dally,
  ``{{EIE}: Efficient Inference Engine on Compressed Deep Neural Network},'' in
  \emph{ISCA 2016 Proc.}\hskip 1em plus 0.5em minus 0.4em\relax {IEEE}, 2016,
  pp. 243--254.

\bibitem{wu2016a}
J.~Wu, C.~Leng, Y.~Wang, Q.~Hu, and J.~Cheng, ``{Quantized Convolutional Neural
  Networks for Mobile Devices},'' in \emph{CVPR 2016 Proc.}\hskip 1em plus
  0.5em minus 0.4em\relax {IEEE}, 2016, pp. 4820--4828.

\bibitem{iandola2016a}
F.~N. Iandola, S.~Han, M.~W. Moskewicz, K.~Ashraf, W.~J. Dally, and K.~Keutzer,
  ``{SqueezeNet: AlexNet-level accuracy with 50x fewer parameters and <0.5MB
  model size},'' \emph{arXiv e-prints}, p. arXiv:1602.07360, 2016.

\bibitem{zoph2018a}
B.~Zoph, V.~Vasudevan, J.~Shlens, and Q.~V. Le, ``{Learning Transferable
  Architectures for Scalable Image Recognition},'' in \emph{CVPR 2018
  Proc.}\hskip 1em plus 0.5em minus 0.4em\relax {IEEE}, 2018, pp. 8697--8710.

\bibitem{zhang2018a}
X.~Zhang, X.~Zhou, M.~Lin, and J.~Sun, ``{{ShuffleNet}: An Extremely Efficient
  Convolutional Neural Network for Mobile Devices},'' in \emph{CVPR 2018
  Proc.}\hskip 1em plus 0.5em minus 0.4em\relax {IEEE}, 2018, pp. 6848--6856.

\bibitem{yang2019a}
J.~Yang, X.~Shen, J.~Xing, X.~Tian, H.~Li, B.~Deng, J.~Huang, and X.-s. Hua,
  ``{Quantization Networks},'' in \emph{{CVPR 2019 Proc.}}\hskip 1em plus 0.5em
  minus 0.4em\relax {IEEE}, 2019, pp. 7300--7308.

\bibitem{zebin2019a}
T.~Zebin, P.~J. Scully, N.~Peek, A.~J. Casson, and K.~B. Ozanyan, ``{Design and
  Implementation of a Convolutional Neural Network on an Edge Computing
  Smartphone for Human Activity Recognition},'' \emph{{IEEE Access}}, vol.~7,
  pp. 133\,509--133\,520, 2019.

\bibitem{nagel2020a}
M.~Nagel, R.~A. Amjad, M.~Van~Baalen, C.~Louizos, and T.~Blankevoort, ``{Up or
  Down? Adaptive Rounding for Post-Training Quantization},'' in \emph{{ICML
  2020 Proc.}}, vol. 119.\hskip 1em plus 0.5em minus 0.4em\relax PMLR, 2020,
  pp. 7197--7206.

\bibitem{cass2019a}
S.~Cass, ``{Taking AI to the edge: Google{\textquotesingle}s {TPU} now comes in
  a maker-friendly package},'' \emph{{IEEE Spectrum}}, vol.~56, no.~5, pp.
  16--17, 2019.

\bibitem{ionica2015a}
M.~H. Ionica and D.~Gregg, ``{The Movidius Myriad
  Architecture{\textquotesingle}s Potential for Scientific Computing},''
  \emph{{{IEEE} Micro}}, vol.~35, no.~1, pp. 6--14, 2015.

\bibitem{xu2017a}
X.~Xu, J.~Amaro, S.~Caulfield, A.~Forembski, G.~Falcao, and D.~Moloney,
  ``{Convolutional neural network on neural compute stick for voxelized
  point-clouds classification},'' in \emph{CISP-BMEI 2017 Proc.}\hskip 1em plus
  0.5em minus 0.4em\relax {IEEE}, 2017, pp. 1--7.

\bibitem{cass2020a}
S.~Cass, ``{Nvidia makes it easy to embed {AI}: The Jetson nano packs a lot of
  machine-learning power into {DIY} projects},'' \emph{{IEEE Spectrum}},
  vol.~57, no.~7, pp. 14--16, 2020.

\bibitem{arlot2010a}
S.~Arlot and A.~Celisse, ``{A survey of cross-validation procedures for model
  selection},'' \emph{{Stat. Surv.}}, vol.~4, pp. 40--79, 2010.

\bibitem{abadi2016a}
M.~Abadi, P.~Barham, J.~Chen, Z.~Chen, A.~Davis, J.~Dean, M.~Devin,
  S.~Ghemawat, G.~Irving, M.~Isard, M.~Kudlur, J.~Levenberg, R.~Monga,
  S.~Moore, D.~G. Murray, B.~Steiner, P.~Tucker, V.~Vasudevan, P.~Warden,
  M.~Wicke, Y.~Yu, and X.~Zheng, ``{TensorFlow: A System for Large-scale
  Machine Learning},'' in \emph{USENIX-OSDI 2016 Proc.}, 2016, pp. 265--283.

\bibitem{ho2019a}
J.~Ho, T.~Tumkaya, S.~Aryal, H.~Choi, and A.~Claridge-Chang, ``{Moving beyond P
  values: data analysis with estimation graphics},'' \emph{Nature Methods},
  vol.~16, no.~7, pp. 565--566, 2019.

\end{thebibliography}

\vfill

\end{document}